\documentclass[journal,11pt,onecolumn]{IEEEtran}

\usepackage{amsmath}
\usepackage{amssymb}
\usepackage{latexsym}
\usepackage{verbatim}
\usepackage{subfigure}
\usepackage{psfrag}
\usepackage{hyperref}

\usepackage{tcolorbox}
\definecolor{darkred}{RGB}{250,0,0}
\definecolor{darkgreen}{RGB}{0,150,0}
\definecolor{myblue}{RGB}{0,0,250}
\definecolor{darkblue}{RGB}{0,0,200}
\hypersetup{colorlinks=true, linkcolor=darkred, citecolor=myblue, urlcolor=darkblue}

\newtheorem{theo}{Theorem}
\newtheorem{prop}{Proposition}
\newtheorem{lema}{Lemma}
\newtheorem{cor}{Corollary}
\newtheorem{rem}{Remark}
\newtheorem{assump}{Assumption}

{\begin{list}               
    {$\bullet$ \hfill}{
        \setlength{\leftmargin}{\parindent}
        \setlength{\parsep}{0.04\baselineskip}
        \setlength{\itemsep}{0.5\parsep}
        \setlength{\labelwidth}{\leftmargin}
        \setlength{\labelsep}{0em}}
    }
{\end{list}}

\providecommand{\cref}[1]{Chapter~\ref{chap:#1}}

\providecommand{\R}{\ensuremath{\mathbb{R}}}

\providecommand{\abs}[1]{\lvert#1\rvert}
\providecommand{\norm}[1]{\lVert#1\rVert}

\providecommand{\set}[1]{\left\{#1\right\}}

\providecommand{\bydef}{\overset{\text{def}}{=}}

\renewcommand{\vec}[1]{\ensuremath{\boldsymbol{#1}}}
\providecommand{\mat}[1]{\ensuremath{\boldsymbol{#1}}}

\providecommand{\mA}{\mat{A}} \providecommand{\mB}{\mat{B}}
\providecommand{\mC}{\mat{C}}

\providecommand{\mI}{\mat{I}}  
  
 \providecommand{\mP}{\mat{P}} 
\providecommand{\mQ}{\mat{Q}} 
  
\providecommand{\mV}{\mat{V}}

\providecommand{\mSigma}{\mat{\Sigma}}
 \providecommand{\mG}{\mat{G}}

\providecommand{\va}{\vec{a}} \providecommand{\vb}{\vec{b}}
\providecommand{\vc}{\vec{c}} 
\providecommand{\vh}{\vec{h}}

 \providecommand{\vs}{\vec{s}}
 \providecommand{\vr}{\vec{r}}
\providecommand{\vg}{\vec{g}}
\providecommand{\vu}{\vec{u}} \providecommand{\vw}{\vec{w}}
\providecommand{\vx}{\vec{x}}

 \providecommand{\vv}{\vec{v}}

\usepackage{lipsum}    
\usepackage{ifthen}
\usepackage{tcolorbox}
\newboolean{showcomments}
\setboolean{showcomments}{true}
\newcommand{\oussama}[1]{\ifthenelse{\boolean{showcomments}}
{ \textcolor{red}{(Oussama says:  #1)}}{}}
\newcommand{\christos}[1]{\ifthenelse{\boolean{showcomments}}
{ \textcolor{blue}{(Christos says: #1)} } {} }
\newcommand{\yue}[1]{\ifthenelse{\boolean{showcomments}}
{ \textcolor{magenta}{(Yue says:  #1)}}{}}

\usepackage{hyperref}

\usepackage{enumitem}

\providecommand{\abs}[1]{\lvert#1\rvert}
\providecommand{\norm}[1]{\lVert#1\rVert}

\providecommand{\set}[1]{\left\{#1\right\}}

\providecommand{\bydef}{\overset{\text{def}}{=}}

\renewcommand{\vec}[1]{\ensuremath{\boldsymbol{#1}}}
\providecommand{\mat}[1]{\ensuremath{\boldsymbol{#1}}}

\providecommand{\mA}{\mat{A}} \providecommand{\mB}{\mat{B}}
\providecommand{\mC}{\mat{C}}

\providecommand{\mI}{\mat{I}}  
  
 \providecommand{\mP}{\mat{P}} 
\providecommand{\mQ}{\mat{Q}} 
  
\providecommand{\mV}{\mat{V}}

\providecommand{\mSigma}{\mat{\Sigma}}
 \providecommand{\mG}{\mat{G}}

\providecommand{\mLm}{\mat{\Lambda}}

\usepackage{balance}
\providecommand{\va}{\vec{a}} \providecommand{\vb}{\vec{b}}
\providecommand{\vc}{\vec{c}} 
\providecommand{\vh}{\vec{h}}

 \providecommand{\vs}{\vec{s}}
 \providecommand{\vr}{\vec{r}}
 
\providecommand{\vg}{\vec{g}}
\providecommand{\vu}{\vec{u}} \providecommand{\vw}{\vec{w}}
\providecommand{\vx}{\vec{x}}

 \providecommand{\vv}{\vec{v}}

\usepackage{algorithm}
\usepackage{algpseudocode}
\algnewcommand\algorithmicforeach{\textbf{Until :}}
\algnewcommand\algorithmicendif{\textbf{End}}
\algnewcommand\ForEach{\item[ \algorithmicforeach]}
\algnewcommand\EndiFF{\item[ \algorithmicendif]}

\usepackage{cite}

\usepackage{color}

\usepackage{enumitem}
\usepackage{commath}

\usepackage{hyperref}

\usepackage{esdiff}

\usepackage{pgf,tikz,psfrag}

\usepackage{dsfont}

\newcommand{\argmin}{\operatornamewithlimits{argmin}}

\providecommand{\vxi}{\vec{\xi}}

\begin{document}

\title{Phase Transitions in Transfer Learning for High-Dimensional Perceptrons}

\author{Oussama Dhifallah and Yue M. Lu
        \thanks{O. Dhifallah and Y. M. Lu are with the John A. Paulson School of Engineering and Applied Sciences, Harvard University, Cambridge, MA 02138, USA (e-mails: oussama$\_$dhifallah@g.harvard.edu,yuelu@seas.harvard.edu).}
\thanks{This research was funded by the Harvard FAS Dean's Fund for Promising Scholarship, and by the US National Science Foundations under grants CCF-1718698 and CCF-1910410.}
}

\maketitle

\begin{abstract}
Transfer learning seeks to improve the generalization performance of a target task by exploiting the knowledge learned from a related source task. Central questions include deciding what information one should transfer and when transfer can be beneficial. The latter question is related to the so-called negative transfer phenomenon, where the transferred source information actually reduces the generalization performance of the target task. This happens when the two tasks are sufficiently dissimilar. In this paper, we present a theoretical analysis of transfer learning by studying a pair of related perceptron learning tasks. Despite the simplicity of our model, it reproduces several key phenomena observed in practice. Specifically, our asymptotic analysis reveals a phase transition from negative transfer to positive transfer as the similarity of the two tasks moves past a well-defined threshold.
\end{abstract}

\section{Introduction}
\label{intro}
Transfer learning \cite{trf1,trf2,tranf_lean,tranf_lean2,tan2018} is a promising approach to improving the performance of machine learning tasks. It does so by exploiting the knowledge gained from a previously-learned model, referred to as the \emph{source task}, to improve the generalization performance of a related learning problem, referred to as the \emph{target task}. One particular challenge in transfer learning is to avoid the so-called \emph{negative transfer} \cite{trsf_no,avd_negt_1,avd_negt_2,kornblith2019better}, where the transferred source information reduces the generalization performance of the target task. Recent literature \cite{trsf_no,avd_negt_1,avd_negt_2,kornblith2019better} shows that negative transfer is closely related to the similarity between the source and target tasks. Transfer learning may hurt the generalization performance if the tasks are sufficiently dissimilar. 

In this paper, we present a theoretical analysis of transfer learning by studying a pair of related perceptron learning tasks. Despite the simplicity of our model, it reproduces several key phenomena observed in practice. Specifically, the model reveals a sharp phase transition from negative transfer to positive transfer (i.e. when transfer becomes helpful) as a function of the model similarity.

\subsection{Models and Learning Formulations}

We start by describing the models for our theoretical study. We assume that the source task has a collection of training data $\lbrace (\va_{s,i},y_{s,i}) \rbrace_{i=1}^{n_s}$, where $\va_{s,i}\in\mathbb{R}^p$ is the source feature vector and $y_{s,i} \in \mathbb{R}$ denotes the label corresponding to $\va_{s,i}$. Following the standard teacher-student paradigm, we shall assume that the labels $\lbrace y_{s,i} \rbrace_{i=1}^{n_s}$ are generated according to the following model
\begin{align}\label{smodel}
y_{s,i}=\varphi(\va_{s,i}^\top \vxi_s),~\forall~i \in \lbrace 1,\dots,n_s \rbrace,
\end{align}
where $\varphi(\cdot)$ is a scalar deterministic or probabilistic function and $\vxi_s\in\mathbb{R}^p$ is an unknown \emph{source teacher vector}.

Similar to the source task, the target task has access to a different collection of training data $\lbrace (\va_{t,i},y_{t,i}) \rbrace_{i=1}^{n_t}$, generated according to 
\begin{align}\label{tmodel}
y_{t,i}=\varphi(\va_{t,i}^\top \vxi_t),~\forall~i \in \lbrace 1,\dots,n_t \rbrace.
\end{align}
Here, $\vxi_t\in\mathbb{R}^p$ is an unknown \emph{target teacher vector}. We measure the (dis)similarity of the two tasks by 
\[
\rho \bydef \frac{\vxi_t^\top \vxi_s}{\norm{\vxi_t} \norm{\vxi_s}},
\]
with $\rho=0$ indicating two uncorrelated tasks, whereas $\rho = 1$ means that the tasks are perfectly aligned.


For the source task, we learn the optimal weight vector $\widehat{\vw}_s$ by solving a convex optimization problem
\begin{align}\label{sform}
\widehat{\vw}_s=\argmin_{\vw \in\mathbb{R}^p}\ \frac{1}{p} \sum_{i=1}^{n_s}  \ell\left(y_{s,i};\va_{s,i}^\top \vw \right)+\frac{\lambda}{2} \norm{\vw}^2.
\end{align}
Here, $\lambda\ge0$ is a regularization parameter, and $\ell(.;.)$ denotes some general loss function that can take one of the following two forms
\begin{align}
\begin{cases}\label{loss_funs}
\ell \left(y;x \right)=\widehat{\ell}(y-x), & \text{for regression task}\\
\ell \left(y;x \right)=\widehat{\ell}(yx), & \text{for classification task},
\end{cases}
\end{align}
where $\widehat{\ell}(.)$ is a convex function.

In this paper, we consider a common strategy in transfer learning \cite{tranf_lean2} which consists of transferring the optimal source vector, i.e. $\widehat{\vw}_s$, to the target task. One popular approach is to fix a (random) subset of the target weights to values of the corresponding optimal weights learned during the source training process \cite{hard_trf1}. In our learning model, this amounts to the following target learning formulation:
\begin{align}\label{tform}
\widehat{\vw}_t=&\argmin_{\vw\in\mathbb{R}^p} \frac{1}{p} \sum_{i=1}^{n_t} \ell \left(y_{t,i};\va_{t,i}^\top \vw \right)+\frac{\lambda}{2} \norm{\vw}^2\\
&~~~\text{s.t.}~~\mQ \vw=\mQ \widehat{\vw}_s.
\end{align}
Here, $\widehat{\vw}_s$ is the optimal solution of the source learning problem, and $\mQ\in\mathbb{R}^{p\times p}$ is a diagonal matrix with diagonal entries drawn independently from a Bernoulli distribution with probability $\delta \leq 1$. Thus, on average, we are retaining $\delta p$ number of entries from the source optimal vector $\widehat{\vw}_s$. In addition to possible improvement to the generalization performance, this approach can considerably lower the computational complexity of the target learning task by reducing the number of free optimization variables. In what follows, we refer to $\delta$ as the \textit{transfer rate} and call \eqref{tform} the \textit{hard transfer} formulation.

Another popular approach in transfer learning is to search for target weight vectors in the vicinity of the optimal source weight vector $\widehat{\vw}_s$. This can be achieved by adding a regularization term to the target formulation \cite{weig_reg,weig_reg2}, which in our model becomes
\begin{align}\label{tform_new}
\widehat{\vw}_t&=\argmin_{\vw\in\mathbb{R}^p} \frac{1}{p} \sum_{i=1}^{n_t} \ell \left(y_{t,i};\va_{t,i}^\top \vw \right)+\frac{\lambda}{2} \norm{\vw}^2+\frac{1}{2} \norm{ \mSigma (\vw - \widehat{\vw}_s) }^2,
\end{align}
with $\mSigma\in\mathbb{R}^{p\times p}$ denoting some weighting matrix. In what follows, we refer to \eqref{tform_new} as the \textit{soft transfer} formulation, since it relaxes the strict equality in \eqref{tform}. In fact, the hard transfer in \eqref{tform} is just a special case of the soft transfer formulation, if we set $\mSigma$ to be a diagonal matrix whose diagonal entries are either $+\infty$ (with probability $\delta$) or $0$ (with probability $1-\delta$). 

To measure the performance of the transfer learning methods, we use the generalization error of the target task. Given a new data sample $(\va_{t,\text{new}}, y_{t, \text{new}})$ with $y_{t,\text{new}}=\varphi(\vxi_t^\top \va_{t,\text{new}})$, we assume that the target task predicts the corresponding label as
\begin{equation}\label{phihat}
\widehat{y}_{t,\text{new}}=\widehat{\varphi} [\widehat{\vw}_t^\top \va_{t,\text{new}}],
\end{equation}
where $\widehat{\varphi}(\cdot)$ is a pre-defined scalar function that might be different from $\varphi(\cdot)$. We then calculate the generalization error of the target task as
\begin{align}\label{gener_trg}
\mathcal{E}_{\text{test}}=\frac{1}{4^\upsilon} \mathbb{E}\left[ \big( y_{t,\text{new}}-\widehat{\varphi}(\widehat{\vw}_t^\top  \va_{t,\text{new}}) \big)^2 \right],
\end{align}
where the expectation is taken with respect to the new data $(\va_{t,\text{new}}, y_{t, \text{new}})$. The variable $\upsilon$ allows us to write a more compact formula: $\upsilon$ is taken to be $0$ for a regression problem and $\upsilon=1$ for a binary classification problem. Finally, we use the training error
\begin{align}
\mathcal{E}_{\text{train}}= \frac{1}{p} \sum_{i=1}^{n_t} \ell \left(y_{t,i};\va_{t,i}^\top \widehat{\vw}_t \right)+\frac{1}{2} \norm{ \mSigma (\widehat{\vw}_t - \widehat{\vw}_s) }^2,\nonumber
\end{align}
to quantify the performance of the training process.


\subsection{Main Contributions}

The main contributions of this paper are two-fold, as summarized below:

\subsubsection{Precise Asymptotic Analysis} We present a precise asymptotic analysis of the transfer learning approaches introduced in \eqref{tform} and \eqref{tform_new} for Gaussian feature vectors. Specifically, we show that, as the dimensions $p, n_s, n_t$ grow to infinity with the ratios $\alpha_s = n_s / p, \alpha_t = n_t / p$ fixed, the generalization errors of the hard and soft formulations can be exactly characterized by the solutions of two low-dimensional \emph{deterministic} optimization problems. (See Theorem~\ref{target_soft} and Corollary~\ref{corollary_hard} for details.) Our asymptotic predictions hold for any convex loss functions used in the training process, including the squared loss for regression problems and logistic loss commonly used for binary classification problems.

\begin{figure}[t]
    \centering
    \subfigure[]{\label{fig_inta}
        \includegraphics[width=0.46\linewidth]{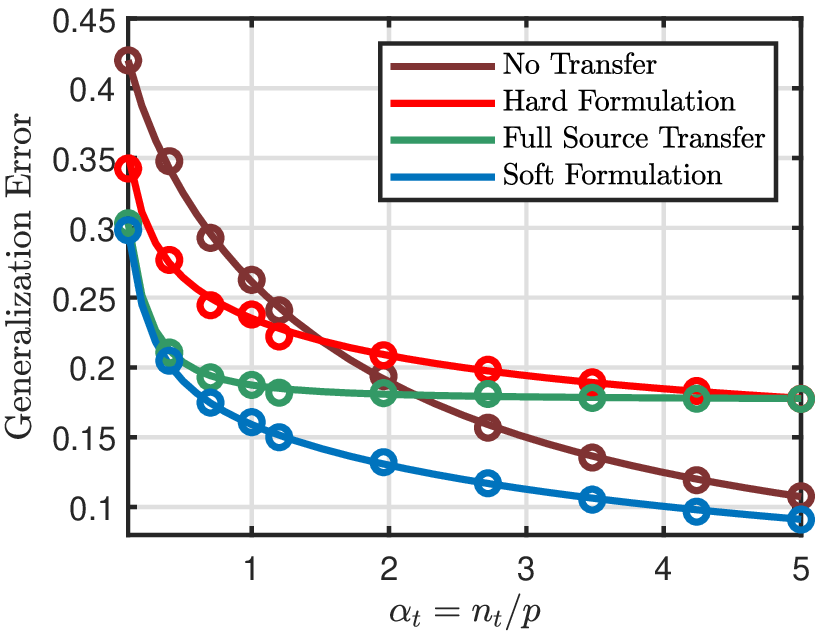}}
        \subfigure[]{\label{fig_intap}
        \includegraphics[width=0.46\linewidth]{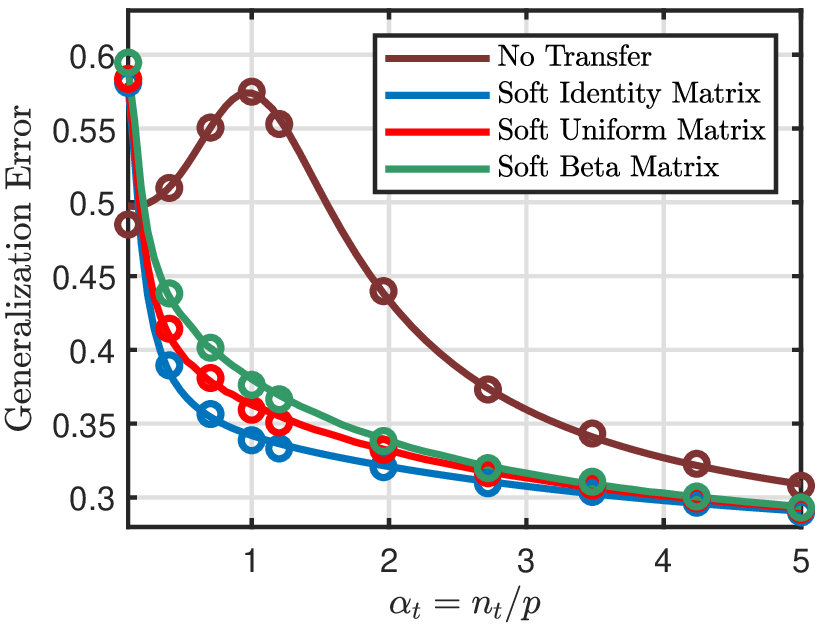}}\\
           \caption{  Theoretical predictions v.s. numerical simulations obtained by averaging over $100$  independent Monte Carlo trials with dimension $p=2500$. {\bf(a)} Binary classification with logistic loss. We take $\alpha_s=10\alpha_t$, $\lambda = 0.3$, $\mSigma=\mI_p/\sqrt{5}$ and $\rho=0.85$, where $\alpha_s=n_s/p$ and $\alpha_t=n_t/p$. The functions $\varphi(\cdot)$ and $\widehat{\varphi}(\cdot)$ are both the sign function. For hard transfer, we set the transfer rate to be $\delta = 0.5$. \emph{Full source transfer} corresponds to $\delta = 1.0$, whereas \emph{no transfer} corresponds to $\delta = 0$. {\bf(b)} Nonlinear regression using quadratic loss, where $\varphi(\cdot)$ is the ReLu function and $\widehat{\varphi}(\cdot)$ is the identity function. Soft identity, beta and uniform matrices refer to different choices of the weighting matrix in \eqref{tform_new}. They correspond to setting $\mSigma$ to be an identity matrix, and a random matrix with diagonal elements drawn from the beta and uniform distributions, respectively. We scale all diagonal elements of $\mSigma$ to have the same mean. We also take $\alpha_s=10\alpha_t$, $\lambda=0.1$, and $\rho=0.8$.}
        \label{fig_int}
\end{figure}

As illustrated in Figure~\ref{fig_int}, our theoretical predictions (drawn as solid lines in the figures) reach excellent agreement with the actual performance (shown as circles) of the transfer learning problem. Figure~\ref{fig_inta} considers a binary classification setting with logistic loss, and we plot the generalization errors of different transfer approaches as a function of the target data/dimension ratio $\alpha_t=n_t/p$. We can see that the hard transfer formulation \eqref{tform} is only useful when $\alpha_t$ is small. In fact, we encounter negative transfer (i.e. hard transfer performing worse than no transfer) when $\alpha_t$ becomes sufficiently large. Moreover, the soft transfer formulation \eqref{tform_new} seems to achieve more favorable generalization errors as compared to the hard formulation. In Figure~\ref{fig_intap}, we consider a regression setting with a squared loss, and explore the impact of different weighting schemes on the performance of the soft formulation. We can see that the soft formulation indeed considerably improves the generalization performance of the standard learning method (i.e. learning the target task without any knowledge transfer). 

\subsubsection{Phase Transitions}
Our asymptotic characterizations reveal a phase transition phenomenon in the hard transfer formulation. Let
\[
\delta^\star = \argmin_{0 \le \delta \le 1} \,\mathcal{E}_{\text{test}}(\delta),
\]
be the optimal transfer rate that minimizes the generalization error of the target task. Clearly, $\delta^\star = 0$ corresponds to the negative transfer regime, where transferring the knowledge of the source task will actually hurt the performance of the target task. In contract, $\delta^\star > 0$ signifies that we have entered the positive transfer regime, where transfer becomes helpful.

\begin{figure}[t]
    \centering
    \subfigure[]{\label{fig_intc}
    \includegraphics[width=0.46\linewidth]{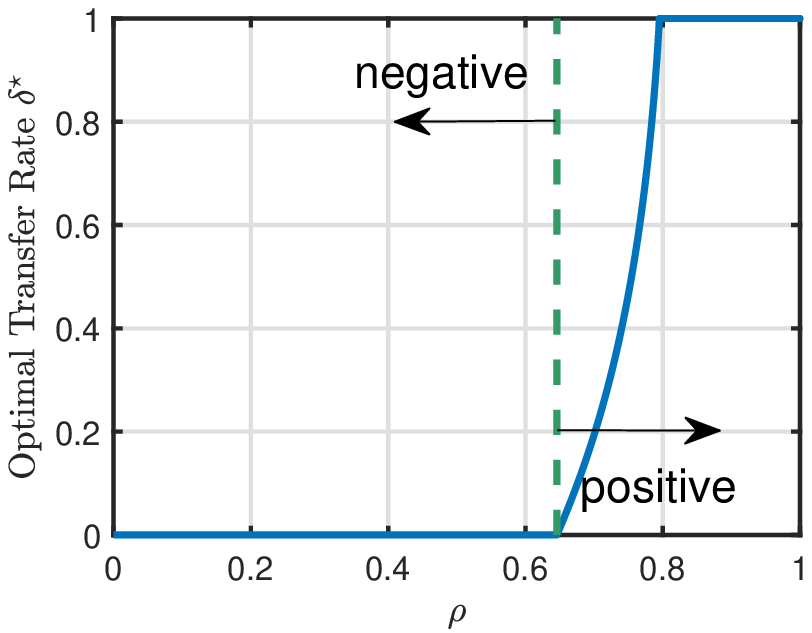}
    }
     \subfigure[]{\label{fig_intb}
    \includegraphics[width=0.46\linewidth]{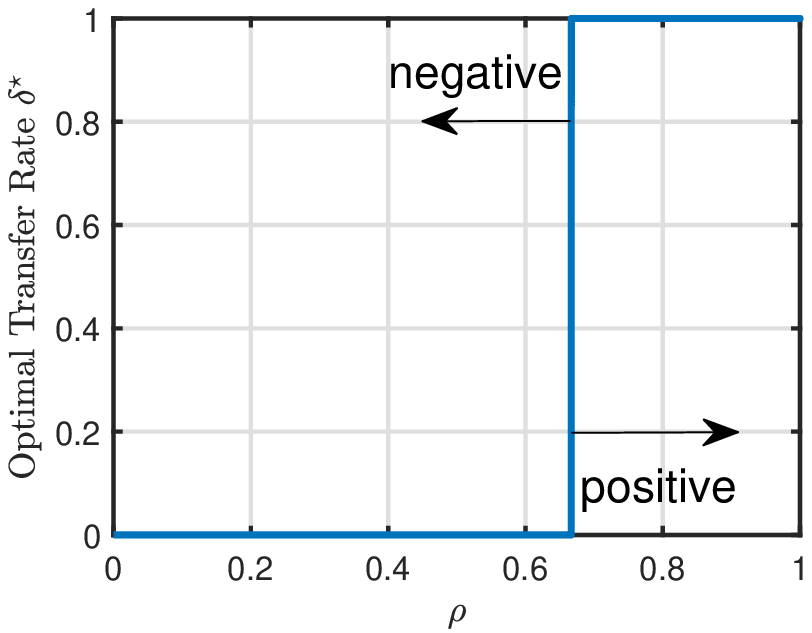}
    }
    \caption{Phase transitions of the hard transfer formulation. When the similarity $\rho$ between the two tasks is small, we are in the negative transfer regime, where we should not transfer the knowledge from the source task. However, as $\rho$ moves past a critical threshold, we enter the positive transfer regime. {\bf(a)} Binary classification with squared loss, with parameters $\alpha_t=2$, $\alpha_s=2\alpha_t$ and $\lambda=0$. Both $\varphi(\cdot)$ and  $\widehat{\varphi}(\cdot)$ are the sign function. {\bf(b)} Nonlinear regression with squared loss, with parameters $\alpha_t=2$, $\alpha_s=2\alpha_t$, and $\lambda=0$. $\varphi(.)$ is the ReLu function and $\widehat{\varphi}(.)$ is the identity function.}
    \label{fig_phase}
\end{figure}

Figure \ref{fig_intc} illustrates the phase transition from negative to positive transfer regimes in a binary classification setting, as the similarity $\rho$ between the two tasks moves past a critical threshold. Similar phase transition phenomena also appear in nonlinear regression, as shown in Figure~\ref{fig_intb}. Interestingly, for this setting the optimal transfer rate \emph{jumps} from $\delta^\star = 0$ to $\delta^\star = 1$ at the transition threshold. 

For general loss functions, the exact locations of the phase transitions can only be found numerically by solving the deterministic optimization problems in our asymptotic characterizations. For the special case of squared loss with no regularization, however, we are able to obtain the following simple analytical characterization for the phase transition threshold: We are in the positive transfer regime \emph{if and only if}
\begin{equation}\label{sharp_trs}
\rho > \rho_c(\alpha_s, \alpha_t) = 1-\frac{\mathbb{E}[\varphi^2(z)]-\mathbb{E}^2[z\varphi(z)]}{2\,\mathbb{E}^2[z\varphi(z)]} \Big(\frac{1}{\alpha_t-1} - \frac{1}{\alpha_s-1}\Big),
\end{equation}
where $z$ is a standard Gaussian random variable. This result is shown in Proposition~\ref{th_phtr}.

By the Cauchy-Schwarz inequality, $\mathbb{E}[\varphi^2(z)]\ge\mathbb{E}^2[z\varphi(z)]$. It follows that $\rho_c(\alpha_s, \alpha_t)$ is an increasing function of $\alpha_t$ and a decreasing function of $\alpha_s$. This property is consistent with our intuition: As we increase $\alpha_t$, the target task has more training data to work with, and thus we should set a higher bar in terms of when to transfer knowledge; As we increase $\alpha_s$, the quality of the optimal source vector becomes better, in which case we can start doing the transfer at a lower similarity level. In particular, when $\alpha_t > \alpha_s$, we have $\rho_c(\alpha_s, \alpha_t) > 1$ and thus the inequality in \eqref{sharp_trs} is never satisfied (because $\abs{\rho} \le 1$ by definition). This indicates that no transfer should be done when the target task has more training data than the source task.


\subsection{Related Work}
The idea of transferring informaton between different domains or different tasks was first proposed in \cite{trf1} and further developed in \cite{trf2}. It has been attracting significant interest in recent literature \cite{tranf_lean2,tan2018, trsf_no,avd_negt_1,avd_negt_2,kornblith2019better,weig_reg,weig_reg2}. While most work focuses on the practical aspects of transfer learning, there have been several studies (e.g., \cite{lampinen2019,rice_trf}) that seek to provide analytical understandings of transfer learning in simplified models. Our work is particularly related to \cite{rice_trf}, which considers a transfer learning model similar to ours, but for the special case of linear regression. The analysis in this paper is more general as it considers arbitrary convex loss functions. We would also like to mention an interesting recent work that studies a different but related setting referred to as knowledge distillation \cite{sag2020solvable}.

In term of technical tools, our asymptotic predictions are derived using the convex Gaussian min-max theorem (CGMT). The CGMT is first introduced in \cite{stojnic2013} and further developed in \cite{chris:151}. It extends a Gaussian comparison inequality first introduced in \cite{gordon}. It particularly uses convexity properties to show the equivalence between two Gaussian processes. The CGMT has been successfully used to analyze convex regression formulations \cite{chris:151,ouss19,dhifallah2020} and convex classification formulations \cite{ea19bc,kam20,mig2020role,aubin2020gener}.

\subsection{Organization}
The rest of this paper is organized as follows. Section \ref{assume_sec} states the technical assumptions under which our results are obtained. Section \ref{prf_sft} provides an asymptotic characterization of the soft transfer formulation. The precise analysis of the hard transfer formulation is presented in Section \ref{hard_analysis}. Our theoretical predictions hold for general convex loss functions. We specialize these results to the settings of nonlinear regression and binary classification in Section \ref{sim_res}, where we also provide additional numerical results to validate our predictions. Section \ref{tech_deta} provides the detailed proof of the technical statements  introduced in Sections \ref{prf_sft} and \ref{hard_analysis}. Section \ref{cond} concludes the paper. The Appendix provides additional technical details.

\section{Technical Assumptions}
\label{assume_sec}
The theoretical analysis of this paper is carried out under the following assumptions.
\begin{assump}[Gaussian Feature Vectors]
\label{rad_fv}
The feature vectors $\lbrace \va_{s,i} \rbrace_{i=1}^{n_s}$ and $\lbrace \va_{t,i} \rbrace_{i=1}^{n_t}$ are drawn independently from a standard Gaussian distribution. The vector $\vxi_s\in\mathbb{R}^p$ can be expressed as $\vxi_s=\rho \vxi_t + \sqrt{1-\rho^2} \vxi_r$, where the vectors ${\vxi}_t\in\mathbb{R}^p$ and ${\vxi}_r\in\mathbb{R}^p$ are independent from the feature vectors, and they are generated independently from a uniform distribution on the unit sphere.
\end{assump}
Define $m$ as the number of transferred entries in \eqref{tform}. Our results are valid in the high-dimensional asymptotic setting where the dimensions $p$, $n_s$, $n_t$ and $m$ grow to infinity at fixed ratios.
\begin{assump}[High-dimensional Asymptotic]
\label{highdim}
The number of samples and the number of transferred components in hard transfer satisfy $n_s=n_s(p)$, $n_t=n_t(p)$ and $m=m(p)$ with $\alpha_{s,p}=n_s(p)/p \to \alpha_s>0$, $\alpha_{t,p}=n_t(p)/p \to \alpha_t>0$ and $\delta_p=m(p)/p \to \delta>0$ as $p\to \infty$.
\end{assump}
\begin{assump}[Loss Function]
\label{lossf}
The loss function ${\ell}(y;.)$ defined in \eqref{loss_funs} is a proper convex function in $\mathbb{R}$. Moreover, define a random function $\mathcal{L}(\vx)=\sum_{i=1}^{n_t} \ell(y_i;x_i)$, where $y_i {\sim}\varphi(z_i)$, with $\set{z_i}$ being a collection of independent standard normal random variables. Denote by $\partial \mathcal{L}$  the sub-differential set of $\mathcal{L}(\vx)$. Then there exists a universal constant ${C}>0$ such that
\[
\mathbb{P}\Big(\sup_{\norm{\vv} \le C \sqrt{n_t}}\ \sup_{\vs\in \partial \mathcal{L}(\vv)} \norm{\vs} \leq {C} \sqrt{n_t}\Big) \xrightarrow{p \to \infty} 1.
\]
\end{assump}
Furthermore, we consider the following assumption to guarantee that the generalization error defined in \eqref{gener_trg} concentrates in the large system limit.
\begin{assump}[Regularity Conditions]
\label{fhat} 
The data generating function $\varphi(.)$ is independent from the feature vectors. Moreover, the following conditions are satisfied.
\begin{itemize}
\item $\varphi(\cdot)$ and $\widehat{\varphi}(\cdot)$ are continuous almost everywhere in $\mathbb{R}$. For every $h > 0$ and $z\sim\mathcal{N}(0,h)$, we have $0<\mathbb{E}[\varphi^2(z)]<+\infty$ and $0<\mathbb{E}[\widehat{\varphi}^2(z)]<+\infty$.
\item For any compact interval $[c, C]$, there exists a function $g(\cdot)$ such that 
\[
\sup_{h \in [c, C]} \abs{ \widehat{\varphi}(h x) }^2 \leq g(x) \quad \text{for all } x \in \R.
\]
Additionally, the function $g(\cdot)$ satisfies $\mathbb{E}[{{g}^2(z)}]<+\infty$, where $z\sim\mathcal{N}(0,1)$.
\end{itemize}
\end{assump}
Finally, we introduce the following assumption to guarantee that the training and generalization errors of the soft formulation can be asymptotically characterized by deterministic optimization problems. 
\begin{assump}[Weighting Matrix]
\label{Sigmaw} 
Let $\mLm = \mSigma^\top \mSigma$ where $\mSigma$ is the weighting matrix in the soft transfer formulation. Let $\sigma_{\text{max}}(\mLm)$ denote its largest eigenvalue, and let $\sigma_{\text{min},1}(\mLm)$ and $\sigma_{\text{min},2}(\mLm)$ denote its two smallest eigenvalues. There exist two constants $\mu_{\text{min}}\geq 0$ and $\mu_{\text{max}}\geq 0$ such that
\begin{align}
\begin{cases}
\mathbb{P}(\sigma_{\text{max}}(\mLm) \leq \mu_{\text{max}})  \xrightarrow{n\to\infty} 1\\
\sigma_{\text{min},1}(\mLm) \xrightarrow{~~p~~} \mu_{\text{min}} \\
\abs{ \sigma_{\text{min},1}(\mLm)-\sigma_{\text{min},2}(\mLm) } \xrightarrow{~~p~~} 0.
\end{cases}
\end{align}
Moreover, we assume that the empirical distribution of the eigenvalues of the matrix $\mLm$ converges weakly to a probability distribution $\mathbb{P}_\mu(.)$ supported in $[\mu_{\text{min}}~\mu_{\text{max}}]$.
\end{assump}

\section{Sharp Asymptotic Analysis of the Soft Transfer Formulation}\label{prf_sft}
In this section, we study the asymptotic properties of the soft transfer formulation. Specifically, we provide a precise characterization of the training and generalization errors corresponding to \eqref{tform_new}. 

The asymptotic performance of the source formulation defined in \eqref{sform} has been studied in the literature \cite{aubin2020gener}. In particular, it has been shown that the asymptotic limit of the source formulation in \eqref{sform} can be quantified by the following deterministic optimization problem:
\begin{align}\label{sc_prob_source}
\min_{\substack{q_s, r_s \geq 0}} \sup_{\substack{\sigma > 0}}& \ \alpha_s \mathbb{E} \Big[\mathcal{M}_{\ell(Y_s,.)}\Big( r_s H_{s}+q_s S_{s}; \frac{r_s}{\sigma}  \Big) \Big] -\frac{r_s \sigma }{2} + \frac{\lambda}{2} (q_s^2+r_s^2).
\end{align}
Here, $Y_s=\varphi(S_{s})$, and $H_s$ and $S_{s}$ are two independent standard Gaussian random variables. Furthermore, the function $\mathcal{M}_{\ell(Y_s,.)}$ introduced in the scalar optimization problem \eqref{sc_prob_source} is the Moreau envelope function defined as
\begin{align}\label{m_env}
\mathcal{M}_{\ell(y,.)}(a;b)=\min_{c\in\mathbb{R}} \ell(y;c)+\frac{1}{2b}(c-a)^2.
\end{align}
The expectation in \eqref{sc_prob_source} is taken over the random variables $H_s$, $S_{s}$.

In our work, we focus on the target problem with soft transfer, as formulated in \eqref{tform_new}. It turns out that the asymptotic performance of the target problem can also be characterized by a deterministic optimization problem:
\begin{align}\label{tg_prob_source}
\min_{\substack{q_t,r_t\geq 0}} &\sup_{\sigma>-\mu_{\text{min}}}    -\frac{\sigma r_t^2}{2} + \frac{1}{2} \left((1-\rho^2) (q_s^\star)^2+(r_s^\star)^2 \right) T_2(\sigma) \nonumber\\
&\hspace{-3mm}+\alpha_t\mathbb{E}\Big[ \mathcal{M}_{\ell(Y_t,.)}\Big( r_t H_t+q_t S_{t}; T_{1}(\sigma) \Big) \Big]+\frac{\lambda}{2} (q_t^2+r_t^2) \nonumber\\
&\hspace{-3mm}-\frac{1}{2} \left(q_t-\rho q_s^\star \right)^2 \left(\sigma-1/T_1(\sigma) \right).
\end{align}
Here, $Y_t=\varphi(S_{t})$, and $H_t$ and $S_{t}$ are independent standard Gaussian random variables. Additionally, $\mu_{\text{min}}$ represents the minimum value of the random variable with distribution $\mathbb{P}_\mu(.)$ as defined in Assumption \ref{Sigmaw}. In the formulation \eqref{tg_prob_source}, the constants $q_s^\star$ and $r_s^\star$ are the optimal solutions of the asymptotic formulation given in \eqref{sc_prob_source}. Moreover, the functions $T_{1}(.)$ and $T_{2}(.)$ are defined as follows:
\begin{align}
T_{1}(\sigma)=\mathbb{E}_\mu[ {1}/{(\mu+\sigma)}],~T_{2}(\sigma)=\mathbb{E}_\mu\left[ {\mu \sigma}/{(\mu+\sigma)} \right],\nonumber
\end{align} 
where the expectations are taken over the probability distribution $\mathbb{P}_\mu(.)$ defined in Assumption \ref{Sigmaw}.

\begin{theo}[Precise Analysis of the Soft Transfer] \label{target_soft}
Suppose that the Assumptions \ref{rad_fv}, \ref{highdim}, \ref{lossf}, \ref{fhat}, and \ref{Sigmaw} are satisfied. Then, the training error corresponding to the soft transfer formulation in \eqref{tform_new} converges in probability as follows 
\begin{align}\label{trg_conv_soft}
\mathcal{E}_{\text{train}} \xrightarrow{p \to \infty} C_t^\star-\frac{\lambda}{2} \big( (q_t^\star)^2+(r_t^\star)^2 \big), 
\end{align}
where $C_t^\star$, $q_t^\star$ and $r_t^\star$ are the optimal objective value and the optimal solution of the scalar formulation in \eqref{tg_prob_source}, respectively. Moreover, the generalization error introduced in \eqref{gener_trg} corresponding to the soft transfer formulation converges in probability as follows 
\begin{align}\label{gen_conv_soft}
{\mathcal{E}}_{\text{test}} \xrightarrow{p \to \infty} \frac{1}{4^\upsilon} \mathbb{E}\left[ \left( \varphi(\nu_1) -\widehat{\varphi}(\nu_2) \right)^2 \right],
\end{align}
where $\nu_1$ and $\nu_2$ are two jointly Gaussian random variables with zero mean and a covariance matrix given by
\begin{align}
\begin{bmatrix}
1 & q_t^\star\\
 q_t^\star &  (q_t^\star)^2+(r_t^\star)^2
\end{bmatrix}.\nonumber
\end{align}
\end{theo}

The proof of Theorem \ref{target_soft} is based on the CGMT framework \cite[Theorem 6.1]{chris:151}. The detailed proof is provided in Section \ref{prd_soft_ana}. The statements in Theorem \ref{target_soft} are valid for a general convex loss function and general learning models that can be expressed as in \eqref{smodel} and \eqref{tmodel}. The analysis in Section \ref{prd_soft_ana} shows that the deterministic problems in \eqref{sc_prob_source} and \eqref{tg_prob_source} are the asymptotic limits of the source and target formulations given in \eqref{sform} and \eqref{tform_new}, respectively. Moreover, it shows that the deterministic problems \eqref{sc_prob_source} and \eqref{tg_prob_source} are strictly convex in the minimization variables and concave in the maximization variables. This implies the uniqueness of the optimal solutions of the minimization problems.
\begin{rem}
The results of the theorem show that the training and generalization errors corresponding to the soft transfer formulation can be fully characterized using the optimal solutions of the scalar formulation in \eqref{tg_prob_source}. Moreover, from its definition, \eqref{tg_prob_source} depends on the optimal solutions of the scalar formulation in \eqref{sc_prob_source} of the source task. This shows that the precise asymptotic performance of the soft transfer formulation can be characterized after solving two scalar deterministic problems.
\end{rem}

\section{Sharp Asymptotic Analysis of Hard Transfer Formulation}\label{hard_analysis}
In this section, we study the asymptotic properties of the hard transfer formulation. We then use these predictions to rigorously prove the existence of phase transitions from negative to positive transfer.

\subsection{Asymptotic Predictions}

As mentioned earlier, the hard transfer formulation can be recovered from \eqref{tform_new} as a special case where the eigenvalues of the matrix $\mLm$ are $+\infty$ with probability $\delta$ and $0$ otherwise. Thus, we obtain the following result as a simple consequence of Theorem \ref{target_soft}.

\begin{cor} \label{corollary_hard}
Suppose that the Assumptions \ref{rad_fv}, \ref{highdim}, \ref{lossf} and \ref{fhat} are satisfied. Then, the asymptotic limit of the hard formulation defined in \eqref{tform} is given by the following deterministic formulation
\begin{align}\label{hard_prob_source}
\min_{\substack{q_t,r_t\geq 0}} \sup_{\sigma>0} \  &\frac{\lambda}{2} (q_t^2+r_t^2) + \frac{\sigma \delta}{2} \big[(1-\rho^2) (q_s^\star)^2+(r_s^\star)^2 \big]  \nonumber\\
&\hspace{-3mm}+\alpha_t\mathbb{E}\Big[ \mathcal{M}_{\ell(Y_t,.)}\Big( r_t H_t+q_t S_{t}; \frac{1-\delta}{\sigma} \Big) \Big]-\frac{\sigma r_t^2}{2} \nonumber\\
&\hspace{-3mm}+\frac{\sigma \delta}{2(1-\delta)} \left(q_t-\rho q_s^\star \right)^2.
\end{align}
Additionally, the training and generalization errors associated with the hard formulation converge in probability to the limits given in \eqref{trg_conv_soft} and \eqref{gen_conv_soft}, respectively.
\end{cor}

\subsection{Phase Transitions}\label{ph_trans_hard}

As illustrated in Figure~\ref{fig_phase}, there is a phase transition phenomenon in the hard transfer formulation, where the problem moves from negative transfer to positive transfer as the similarity of the source and target tasks increases. For general loss functions, the exact location of the phase transition boundary can only be determined by numerically solving the scalar optimization problem in \eqref{hard_prob_source}. 

For the special case of squared loss, however, we are able to obtain analytical expressions. For the rest of this section, we restrict our discussions to the following special settings:
\begin{itemize}
\item[(a)] The loss function $\ell(\cdot,\cdot)$ in \eqref{sform} and \eqref{tform} is the squared loss, i.e. ${\ell}(y,x)=\frac{1}{2}(y-x)^2$.
\item[(b)] The regularization strength $\lambda=0$ in the source and target formulations \eqref{sform} and \eqref{tform}.
\item[(c)] The data/dimension ratios $\alpha_s$ and $\alpha_t$ satisfy $\alpha_s>1$ and $\alpha_t>1$.
\end{itemize}


We first consider a nonlinear regression task, where the function $\varphi(\cdot)$ in the generative models \eqref{smodel} and \eqref{tmodel} can be arbitrary, and the function $\widehat{\varphi}(\cdot)$ in \eqref{phihat} is the identity function.

\begin{prop}[Regression Phase Transition]
\label{th_phtr}
In addition to the conditions (a)--(c) introduced above, assume that the pre-defined function $\widehat{\varphi}(\cdot)$ in \eqref{phihat} is the identity function. Let $\delta^\star$ be the optimal transfer rate that leads to the lowest generalization error in the hard formulation \eqref{tform}. Then, 
\begin{align}\label{asy_delta1}
\delta^\star = \begin{cases}
0 & \text{if}~ \rho < \rho_c(\alpha_s,\alpha_t)\\
1 & \text{if}~ \rho > \rho_c(\alpha_s,\alpha_t),
\end{cases}
\end{align}
where $\rho_c(\alpha_s,\alpha_t)$ is defined in \eqref{sharp_trs}.
\end{prop}

The result of Proposition \ref{th_phtr}, whose proof can be found in Section \ref{reg_pt_pf}, shows that $\rho_c(\alpha_s, \alpha_t)$ is the phase transition boundary separating the negative transfer regime from the positive transfer regime. When the similarity metric $\rho < \rho_c(\alpha_s, \alpha_t)$, the optimal transfer ratio $\delta^\star = 0$, indicating that we should not transfer any source knowledge. Transfer becomes helpful only when $\rho$ moves past the threshold. Note that for this particular model, there is also an interesting feature that the optimal $\delta^\star$ jumps to 1 in the positive transfer phase, meaning that we should fully copy the source weight vector. 

Next, we consider a binary classification task, where the nonlinear functions $\varphi(\cdot)$ and $\widehat{\varphi}(\cdot)$ are both the sign function. We first define a function
\begin{align}
g(\alpha_t,\alpha_s)=1-\frac{(1-\frac{2}{\pi})\alpha_t(\alpha_s-\alpha_t)}{(\alpha_s-1)\left[ \frac{4}{\pi}(\alpha_t-1)\alpha_t+2(1-\frac{2}{\pi})(\alpha_t-1) \right]}.
\end{align}

\begin{prop}[Classification]
\label{th_phtr_class}
Assume that the conditions (a)-(c) introduced above hold, and both $\varphi(\cdot)$ and $\widehat{\varphi}(\cdot)$ are the sign function. Then
\begin{align}\label{asy_delta_class}
\delta^\star > 0 \quad \text{if} \quad \rho > g(\alpha_t, \alpha_s).
\end{align}
\end{prop}

We prove this result at the end of Section \ref{tech_deta}. Unlike \eqref{asy_delta1}, the result in \eqref{asy_delta_class} only provides a \emph{sufficient} condition for when the hard transfer is beneficial. Nevertheless, our numerical simulations show that the sufficient condition in \eqref{asy_delta_class} is actually the correct phase transition boundary for the majority of parameter settings for $\alpha_t, \alpha_s$.


\section{Additional Simulation Results}
\label{sim_res}

In this section, we provide additional simulation examples to confirm our asymptotic analysis and illustrate the phase transition phenomenon. In our experiments, we focus on the regression and classification models.

\subsection{Model Assumptions}
For the regression model, we assume that the source, target and test data are generated according to
\begin{align}\label{relu_mod}
y_i=\max(\va_i^\top \vxi,0),~\forall i \in\lbrace 1,\dots,n \rbrace.
\end{align}
The data $\lbrace (\va_i,y_i) \rbrace_{i=1}^{n}$ can be the training data of the source or target tasks. In this regression model, we assume that the function $\widehat{\varphi}(.)$ is the identity function, i.e. $\widehat{\varphi}(x)=x$. Then, the generalization error corresponding to the soft formulation converges in probability as follows
\begin{align}
\mathcal{E}_{\text{test}} \xrightarrow{p\to \infty} v-2 c q_{t}^\star +( (q_{t}^\star)^2+(r_{t}^\star)^2 ),\nonumber
\end{align}
where $c$ and $v$  are defined as follows
\begin{align}
&c=\mathbb{E}[z\max(z,0)],~v=\mathbb{E}[\max(z,0)^2].\nonumber
\end{align} 
Here, $z$ is a standard Gaussian random variable and $q_{t}^\star$ and $r_{t}^\star$ are defined in Theorem \ref{target_soft}. Additionally, the asymptotic limit of the generalization error corresponding to the hard formulation can be expressed in a similar fashion.

For the binary classification model, we assume that the source, target and test data labels are binary and generated as follows:
\begin{align}\label{bin_mod}
y_i=
\text{sign}(\va_i^\top \vxi),~\forall i \in\lbrace 1,\dots,n \rbrace.
\end{align}
Here, the data $\lbrace (\va_i,y_i) \rbrace_{i=1}^{n}$ can be the training data of the source and target tasks. In this classification model, the objective is to predict the correct sign of any unseen sample $y_{\text{new}}$. Then, we fix the function $\widehat{\varphi}(.)$ to be the sign function. Following Theorem \ref{target_soft}, it can be easily shown that the generalization error corresponding to the soft formulation given in \eqref{tform_new} converges in probability as follows 
\begin{align}
\mathcal{E}_{\text{test}} \xrightarrow{p \to \infty} \frac{1}{\pi} \cos^{-1}\Big( \frac{q_t^\star}{ \sqrt{(q_t^\star)^2+(r_t^\star)^2 } }  \Big).\nonumber
\end{align}
Here, $q_t^\star$ and $r_t^\star$ are the optimal solutions of the target scalar formulation given in \eqref{tg_prob_source}. The generalization error corresponding to the hard formulation given in \eqref{tform} can be expressed in a similar fashion.

\subsection{Phase Transitions in the Hard Formulation}

\begin{figure}[t!]
    \centering
    \subfigure[]{\label{fig_sim1ap}
    \includegraphics[width=0.46\linewidth]{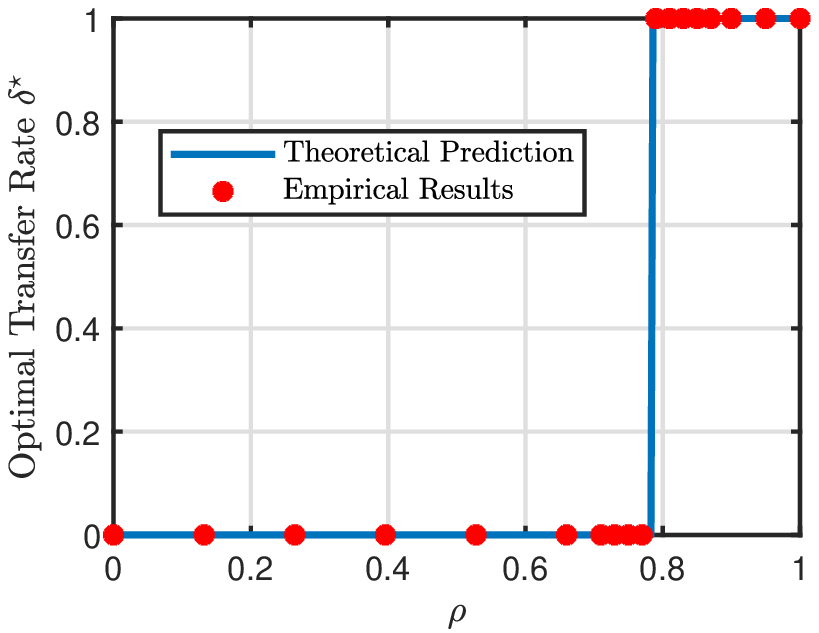}
    }
    \subfigure[]{\label{fig_sim1a}
    \includegraphics[width=0.46\linewidth]{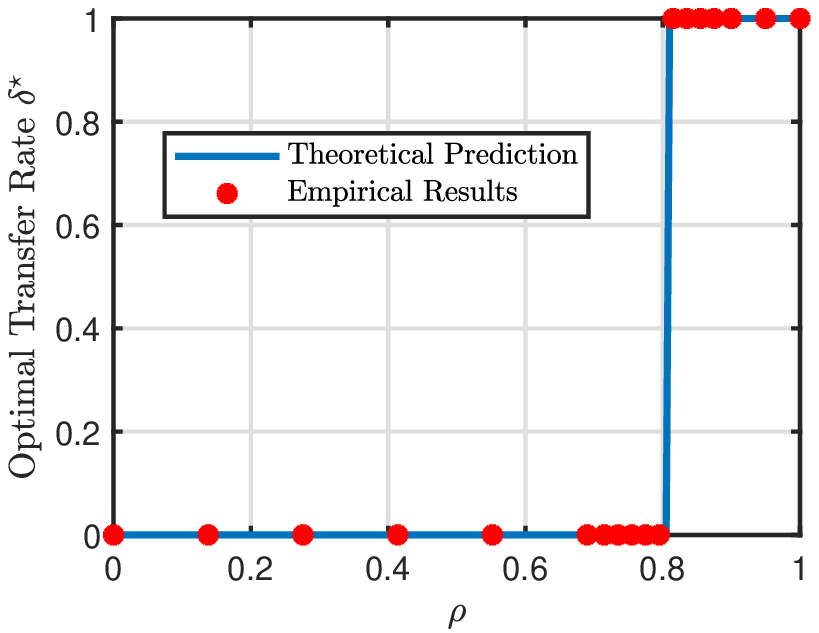}
    }\\
    \subfigure[]{\label{fig_sim1b}
    \includegraphics[width=0.46\linewidth]{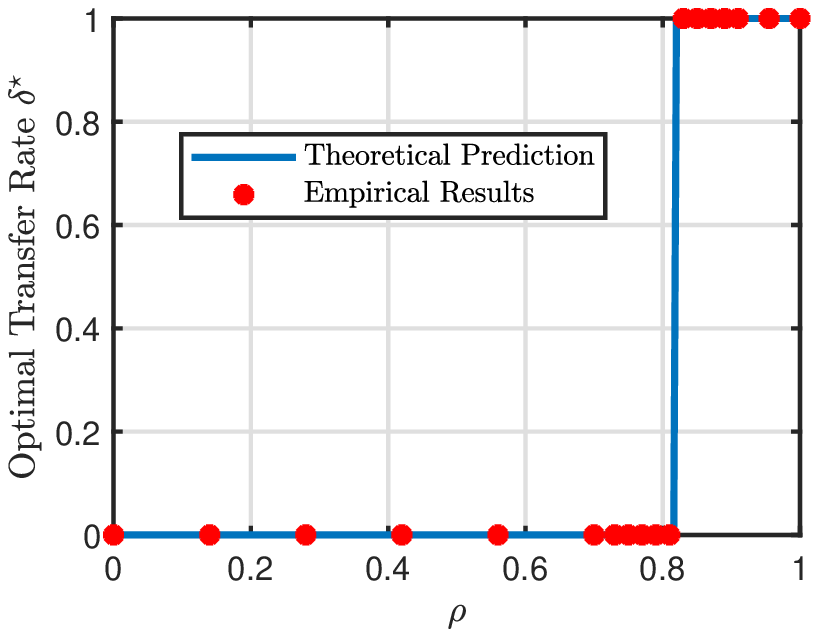}
    }
    \subfigure[]{\label{fig_sim1bp}
    \includegraphics[width=0.46\linewidth]{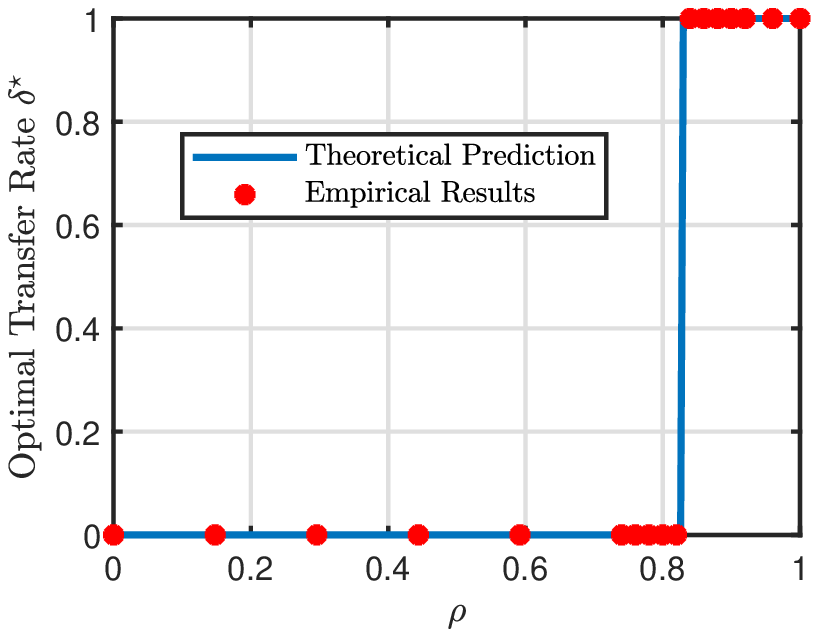}
    }
    \caption{Additional illustrations of the phase transition phenomenon. {\bf(a)} Regression (squared loss, $\alpha_t=0.5$, and $\alpha_s=3\alpha_t$) {\bf(b)} Regression (squared loss, $\alpha_t=2$, and $\alpha_s=2\alpha_t$) {\bf(c)} Binary classification (squared loss, $\alpha_t=1.5$, and $\alpha_s=3\alpha_t$) {\bf(d)} Binary classification (hinge loss, $\alpha_t=1.5$, and $\alpha_s=3\alpha_t$). In all the experiments, we set the regularization strength to be $\lambda=0.1$. The blue line represents our theoretical predictions of the optimal transfer rate obtained by solving our asymptotic results in Section \ref{hard_analysis} for multiple values of $\delta$. The empirical results are averaged over $100$ independent Monte Carlo trials with $p=2500$.}
        \label{fig_sim1}
\end{figure}

In Section~\ref{hard_analysis}, we have presented analytical formulas for the phase transition phenomenon, but only for the special case of squared loss with no regularization. The main purpose of this experiment, shown in Figure \ref{fig_sim1}, is to demonstrate that the phase transition phenomenon still takes place in more general settings with different loss functions and regularization strengths.

In all the cases shown in Figure \ref{fig_sim1}, the transition from negative to positive transfer is a discontinuous jump from standard learning (i.e. no transfer) to full source transfer. Additionally, Figures \ref{fig_sim1b} and \ref{fig_sim1bp} show that the loss function has a small effect on the phase transition boundary. 

\subsection{Soft Transfer: Impact of the Weighting Matrix and Regularization Strength}

In this experiment, we empirically explore the impact of the weighting matrix $\mSigma$ on the generalization error corresponding to the soft formulation. We focus on the binary classification problem with logistic loss. The weighting matrix in \eqref{tform_new} takes the following form
\begin{equation}\label{beta}
\mSigma = \sqrt{\beta_t} \mV,
\end{equation}
where $\mV$ is a diagonal matrix generated in three different ways. (1) \emph{Soft Identity}: $\mV$ is an identity matrix; (2) \emph{Soft Uniform}: the diagonal entries of $\mV$ are drawn independently from the uniform distribution and then scaled to have their mean equal to 1; (3): \emph{Soft Beta}: similar to (2), but with the diagonal entries drawn from the beta distribution, followed by rescaling to unit mean.

\begin{figure}[t!]
    \centering
    \subfigure[]{\label{fig_sim3a}
    \includegraphics[width=0.4725\linewidth]{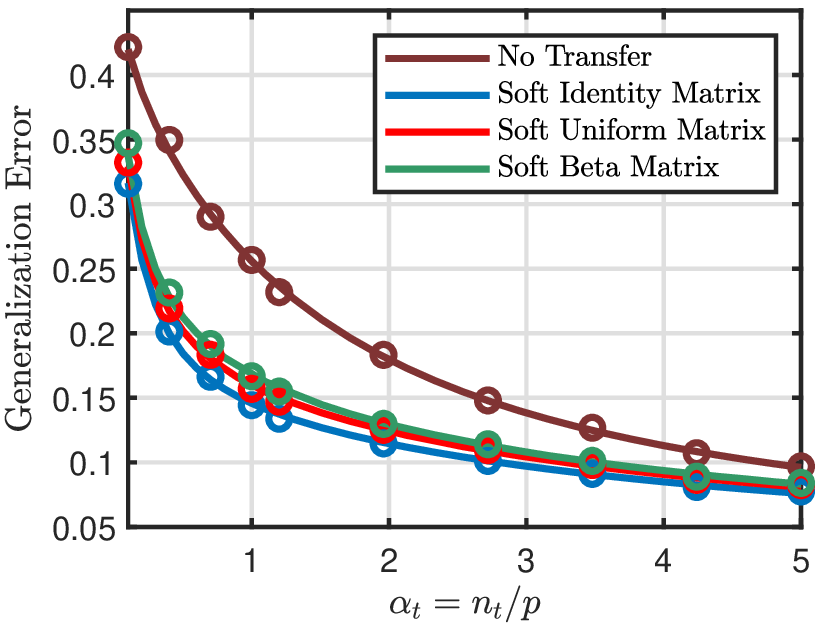}
    }
    \subfigure[]{\label{fig_sim3b}
        \includegraphics[width=0.4725\linewidth]{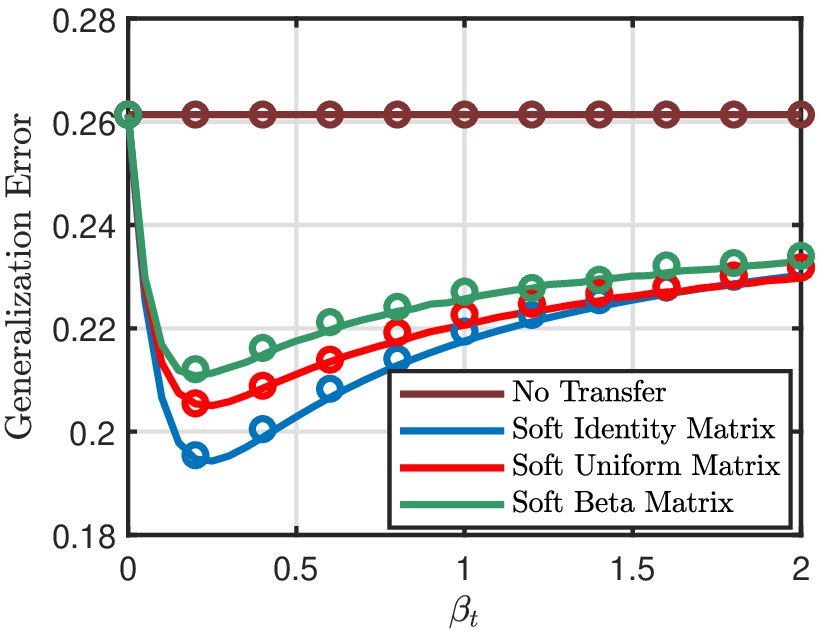}}

    \caption{Continuous line: Theoretical predictions. Circles: numerical simulations. {\bf(a)} $\alpha_s=6\alpha_t$, $\lambda=0.1$, $\beta_t=1/10$ and $\rho=0.9$. {\bf(b)} $\alpha_t=1$, $\alpha_s=5\alpha_t$, $\lambda=0.3$ and $\rho=0.75$.  In all the experiments, we consider the binary classification problem with the logistic loss function. The empirical results are averaged over $50$ independent Monte Carlo trials and we set $p=1000$.}
        \label{fig_sim3}
\end{figure}

Figure \ref{fig_sim3a} shows that the considered weighting matrix choices have similar generalization performance, with the identity matrix being slightly better than the other alternatives. Moreover, Figure \ref{fig_sim3b} illustrates the effects of the parameter $\beta_t$ in \eqref{beta} on the generalization performance. It points to the interesting possibility of ``designing'' the optimal weight matrix to minimize the generalization error.

\subsection{Soft and Hard Transfer Comparison}

In the last simulation example, we consider the regression model and compare the performance of the hard and soft transfer formulations as a function of $\alpha_t$ and $\rho$. 
\begin{figure}[t!]
    \centering
    \subfigure[]{\label{fig_sim2a}
    \includegraphics[width=0.4725\linewidth]{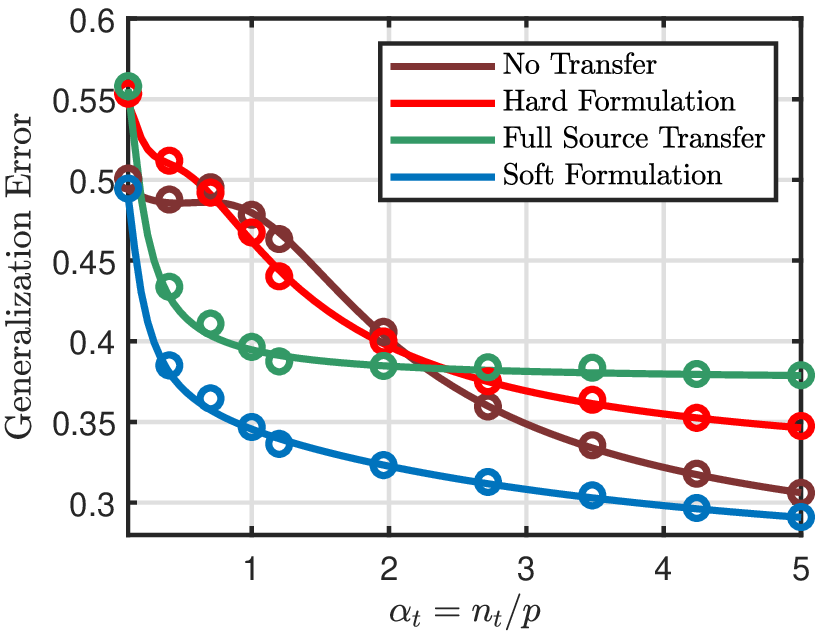}
    }
    \subfigure[]{\label{fig_sim2b}
        \includegraphics[width=0.4725\linewidth]{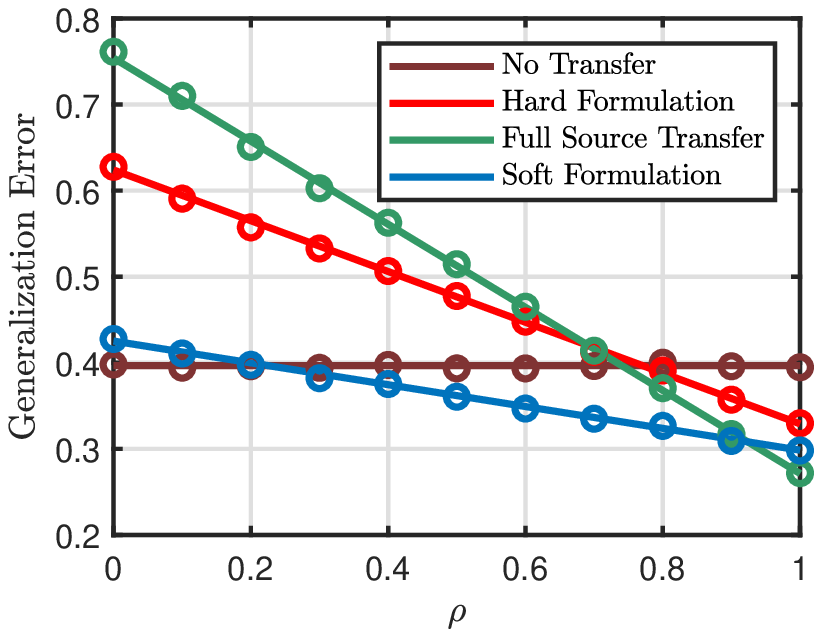}}

    \caption{Continuous line: Theoretical predictions. Circles: numerical simulations. {\bf(a)} $\alpha_s=12\alpha_t$, $\lambda=0.2$ and $\rho=0.75$. {\bf(b)} $\alpha_t=1.5$, $\alpha_s=8\alpha_t$ and $\lambda=0.4$. In all the experiments, we consider the regression setting with a squared loss. The hard transfer formulation uses $\delta=0.5$, and the soft transfer formulation uses an identity weighting matrix. The empirical results are averaged over $50$ independent Monte Carlo trials and we set $p=1000$.}
        \label{fig_sim2}
\end{figure}

Figure \ref{fig_sim2a} shows that the soft formulation provides the best generalization performance for all values of $\alpha_t$. Moreover, we can see that the hard transfer formulation is only useful for small values $\alpha_t$. Figure \ref{fig_sim2b} shows that the performance of the soft and hard transfer formulations depend on the similarity between the source and target tasks. Specifically, the generalization performances of different transfer approaches all improve as we increase the similarity measure $\rho$. We can also see that the full source transfer approach provides the lowest generalization error when the similarity measure is close to $1$, while the soft transfer method leads to the best generalization performance at moderate values of the similarity measure. At very small values of $\rho$, which means that the two tasks share little resemblance, the standard learning method (i.e. no transfer) is the best scheme one should use.

\section{Technical Details}\label{tech_deta}
In this section, we provide a detailed proof of Theorem \ref{target_soft}, Corollary \ref{corollary_hard} and Proportions \ref{th_phtr} and \ref{th_phtr_class}. Specifically, we focus on analyzing the generalized formulation in \eqref{tform_new} using the CGMT framework introduced in the following part.
\subsection{Technical Tool: Convex Gaussian Min-Max Theorem}\label{CGMT_fram}
The CGMT provides an asymptotic equivalent formulation of primary optimization (PO) problems of the following form
\begin{equation}\label{eq:PO}
\Phi_p(\mG)=\min\limits_{\vw \in \mathcal{S}_{\vw}} \max\limits_{\vu\in\mathcal{S}_{\vu}} \vu^\top \mG \vw + \psi(\vw,\vu).
\end{equation}
Specifically, the CGMT shows that the PO given in \eqref{eq:PO} is asymptotically equivalent to the following formulation
\begin{equation}\label{eq:AO}
\hspace{-1mm}\phi_p(\vg,\vh)=\min\limits_{\vw \in \mathcal{S}_{\vw}} \max\limits_{\vu\in\mathcal{S}_{\vu}}  \norm{\vu} \vg^\top \vw +\norm{\vw} \vh^\top \vu  + \psi(\vw,\vu),\nonumber
\end{equation}
referred to as the auxiliary optimization (AO) problem.
Before showing the equivalence between the PO and AO, the CGMT assumes that $\mG\in\mathbb{R}^{{n}\times {p}}$, $\vg \in \mathbb{R}^{{p}}$ and $\vh\in\mathbb{R}^{{n}}$, all have i.i.d standard normal entries, the feasibility sets $\mathcal{S}_{\vw}\subset\R^{{p}}$ and $\mathcal{S}_{\vu}\subset\R^{{n}}$ are convex and compact, and the function $\psi(.,.): \mathbb{R}^{{p}} \times \mathbb{R}^{{n}} \to \mathbb{R}$ is continuous \emph{convex-concave} on $\mathcal{S}_{\vw}\times \mathcal{S}_{\vu}$. Moreover, the function $\psi(.,.)$ is independent of the matrix $\mG$. Under these assumptions, the CGMT \cite[Theorem 6.1]{chris:151} shows that for any $\chi \in \mathbb{R}$ and $\zeta>0$, it holds
\begin{equation}\label{eq:cgmt}
\mathbb{P}\left( \abs{\Phi_p(\mB)-\chi} > \zeta \right) \leq 2 \mathbb{P}\left(  \abs{\phi_p(\vg,\vh)-\chi} > \zeta \right).
\end{equation}
Additionally, the CGMT \cite[Theorem 6.1]{chris:151} provides the following conditions under which the optimal solutions of the PO and AO concentrates around the same set.
\begin{theo}[CGMT Framework]\label{mcgmt}
Consider an open set $\mathcal{S}_{p,\epsilon}$. Moreover, define the set $\mathcal{S}^c_{p,\epsilon}={\mathcal{S}}_{\vw} \setminus \mathcal{S}_{p,\epsilon}$. Let $\phi_p$ and $\phi^c_{p}$ be the optimal cost values of the AO formulation in \eqref{eq:AO} with feasibility sets ${\mathcal{S}}_{\vw}$ and $\mathcal{S}^c_{p,\epsilon}$, respectively. Assume that the following properties are all satisfied
\begin{itemize}
\item[(1)] There exists a constant $\phi$ such that the optimal cost $\phi_p$ converges in probability to $\phi$ as $p$ goes to $+\infty$.
\item[(2)] There exists a positive constant $\zeta>0$ such that $\phi^c_{p} \geq \phi+\zeta$ with probability going to $1$ as $p\to+\infty$, for any fixed $\epsilon>0$.
\end{itemize}
Then, the following convergence in probability holds
\begin{equation}
\abs{ \Phi_p -\phi_p } \overset{p}{\longrightarrow} 0,~\text{and}~\mathbb{P}( \widehat{\vw}_{p} \in \mathcal{S}_{p,\epsilon} )  \overset{p\to\infty}{\longrightarrow} 1,\nonumber
\end{equation}
for any fixed $\epsilon>0$, where $\Phi_p$ and $\widehat{\vw}_{p}$ are the optimal cost and the optimal solution of the PO formulation in \eqref{eq:PO}.
\end{theo} 
Theorem \ref{mcgmt} allows us to analyze the generally easy AO problem to infer asymptotic properties of the generally hard PO problem. Next, we use the CGMT to rigorously prove the technical results presented in Theorem \ref{target_soft}. 

\subsection{Precise Analysis of the Source Formulation}\label{sour_form}
The source formulation defined in \eqref{sform} is well--studied in recent literature \cite{pmlrThramp15}. Specifically, it has been rigorously proved that the performance of the source formulation can be fully characterized after solving the following scalar formulation
\begin{align}\label{sour_prob_pr}
\min_{\substack{q_s, r_s \geq 0}} \sup_{\substack{\sigma > 0}}& \ \alpha_s \mathbb{E} \Big[\mathcal{M}_{\ell(Y_s,.)}\Big( r_s H_{s}+q_s S_{s}; \frac{r_s}{\sigma}  \Big) \Big]  \nonumber\\
&-\frac{r_s \sigma }{2} + \frac{\lambda}{2} (q_s^2+r_s^2).
\end{align}
Here, $Y_s=\varphi(S_{s})$, and $H_s$ and $S_{s}$ are two independent standard Gaussian random variables. The expectation in \eqref{sour_prob_pr} is taken over the random variables $H_s$ and $S_{s}$. Furthermore, the function $\mathcal{M}_{\ell(Y_s,.)}$ introduced in the scalar optimization problem \eqref{sour_prob_pr} is the Moreau envelope function defined in \eqref{m_env}. 

\subsection{Precise Analysis of the Soft Transfer Approach}\label{prd_soft_ana}
In this part, we provide a precise asymptotic analysis of the generalized transfer formulation given in \eqref{tform_new}. Specifically, we focus on analyzing the following formulation 
\begin{align}
\min_{\vw\in\mathbb{R}^p}& \frac{1}{p} \sum_{i=1}^{n_t} \ell \left(y_i;\va_i^\top \vw \right)+\frac{\lambda}{2} \norm{\vw}^2+\frac{1}{2} \norm{ \mSigma (\vw - \widehat{\vw}_s) }^2,\nonumber
\end{align}
where $\widehat{\vw}_s$ is the optimal solution of the source formulation given in \eqref{sform}. Note that the vector $\widehat{\vw}_s$ is independent of the training data of the target task. For simplicity of notation, we denote by $\lbrace (\va_i,y_i) \rbrace_{i=1}^{n_t}$, the training data of the target task. Here, we use the CGMT framework introduced in Section \ref{CGMT_fram} to precisely analyze the above formulation.

\subsubsection{Formulating the Auxiliary Optimization Problem}\label{form_ao}
Our first objective is to rewrite the generalized formulation in the form of the PO problem given in \eqref{eq:PO}. To this end, we introduce additional optimization variables. Specifically, the generalized formulation can be equivalently formulated as follows
\begin{align}\label{po_form_v2}
\min_{\vw\in\mathbb{R}^p} \max_{ \vu \in \mathbb{R}^{n_t}}&~\frac{1}{p}\vu^\top \mA \vw-\frac{1}{p} \sum_{i=1}^{n_t} \ell^\star \left(y_i;u_i \right)+\frac{\lambda}{2} \norm{\vw}^2\nonumber\\
&+\frac{1}{2} \norm{ \mSigma (\vw - \widehat{\vw}_s) }^2.
\end{align}
Here, the optimization vector $\vu\in\mathbb{R}^{n_t}$ is formed as $\vu=[u_{1},\dots,u_{n_t}]^\top$, the data matrix $\mA\in\mathbb{R}^{n_t \times p}$ is given by $\mA=[\va_1,\dots,\va_m]^\top$. Additionally, the function $\ell^\star(y;.)$ denotes the convex conjugate function of the loss function $\ell(y;.)$. First, observe that the CGMT framework assumes that the feasibility sets of the minimization and maximization problems are compact. Then, our next step is to show that the formulation given in \eqref{po_form_v2} satisfies this assumption.
\begin{lema}[Primal--Dual Compactness]\label{lem_compact}
Assume that $\widehat{\vw}$ and $\widehat{\vu}$ are the optimal solutions of the optimization problem in \eqref{po_form_v2}. Then, there exist two constants $C_w>0$ and $C_u>0$ such that the following convergence in probability holds
\begin{align}
\mathbb{P}( \norm{\widehat{\vw}} \leq C_w) \xrightarrow{~p~} 1,~\mathbb{P}( \norm{\widehat{\vu}}/\sqrt{n_t} \leq C_u) \xrightarrow{~p~} 1.
\end{align}
\end{lema}
The proof of Lemma \ref{lem_compact} is omitted since it follows the same steps of the results presented in \cite[Lemma 1]{dhifallah2020} and \cite[Lemma 2]{dhifallah2020}. The proof of the above result follows using Assumption \ref{lossf}, Assumption \ref{Sigmaw} and the asymptotic results in \cite[Theorem 2.1]{rudelson2010} to prove the compactness of the optimal solution $\widehat{\vw}$. Then, use the result in \cite[Proposition 11.3]{var_prog} and Assumption \ref{lossf} to show the compactness of the optimal dual vector $\widehat{\vu}$.

The theoretical result in Lemma \ref{lem_compact} shows that the optimization problem in \eqref{po_form_v2} can be equivalently formulated with compact feasibility sets on events with probability going to one. Then, it suffices to study the constrained version of \eqref{po_form_v2}. Note that the data labels $\lbrace y_i \rbrace_{i=1}^{n_t}$ depend on the data matrix $\mA$. Then, one can decompose the matrix $\mA$ as follows
\begin{align}
\mA&=\mA \mP_{\vxi_t}+\mA \mP^{\perp}_{\vxi}=\mA {\vxi}_t {\vxi}_t^\top+\mA \mP^{\perp}_{\vxi}.\nonumber
\end{align}
Here, the matrix $\mP_{\vxi_t}\in\mathbb{R}^{p\times p}$ denotes the projection matrix onto the space spanned by the vector $\vxi_t$, and the matrix $\mP^{\perp}_{\vxi}=\mI_p-{\vxi}_t {\vxi}_t^\top$ denotes the projection matrix onto the orthogonal complement of the space spanned by the vector $\vxi_t$. Note that we can express $\mA$ as follows without changing its statistics
\begin{align}
\mA=\vs_t {\vxi}_t^\top + \mG \mP^{\perp}_{\vxi},
\end{align}
where $\vs_t \sim \mathcal{N}(0,\mI_{n_t})$ and the components of the matrix $\mG\in\mathbb{R}^{n_t\times p}$ are drawn independently from a standard Gaussian distribution and where $\vs_t$ and $\mG$ are independent. This means that the formulation in \eqref{po_form_v2} can be expressed as follows
\begin{align}\label{po_form_v3}
&\hspace{-2mm}\min_{\norm{\vw}\leq C_w} \max_{\vu\in\mathcal{C}_t}\frac{1}{p}\vu^\top \mG \mP^{\perp}_{\vxi} \vw+\frac{1}{p}\vu^\top \vs_t {\vxi}_t^\top \vw+\frac{\lambda}{2} \norm{\vw}^2\nonumber\\
&-\frac{1}{p} \sum_{i=1}^{n_t} \ell^\star \left(y_i;u_i \right)+\frac{1}{2} \norm{ \mSigma (\vw - \widehat{\vw}_s) }^2,
\end{align}
where the set $\mathcal{C}_t=\lbrace \vu: {\norm{\vu}}/{\sqrt{n_t}}\leq C_u \rbrace$. Note that the formulation in \eqref{po_form_v3} is in the form of the primary formulation given in \eqref{eq:PO}. Here, the function $\psi(.,.)$ is defined as follows
\begin{align}
\psi(\vw,\vu)&=\frac{1}{p}\vu^\top \vs_t {\vxi}_t^\top \vw+\frac{\lambda}{2} \norm{\vw}^2-\frac{1}{p} \sum_{i=1}^{n_t} \ell^\star \left(y_i;u_i \right)\nonumber\\
&+\frac{1}{2} \norm{ \mSigma (\vw - \widehat{\vw}_s) }^2.
\end{align}
One can easily see that the optimization problem in \eqref{po_form_v3} has compact convex feasibility sets. Moreover, the function $\psi(.,.)$ is continuous, convex--concave and independent of the Gaussian matrix $\mG$. This shows that the assumptions of the CGMT are all satisfied by the primary formulation in \eqref{po_form_v3}. Then, following the CGMT framework, the auxiliary formulation corresponding to our primary problem in \eqref{po_form_v3} can be expressed as follows
\begin{align}\label{ao_form_f}
&\hspace{-2mm}\min_{\norm{\vw}\leq C_w} \max_{\vu\in\mathcal{C}_t} \frac{\norm{\vu}}{p}  \vg^\top \mP^{\perp}_{\vxi} \vw+ \frac{1}{p} \vu^\top \vs_t {\vxi}_t^\top \vw + \frac{\vh^\top \vu}{p}\norm{\mP^{\perp}_{\vxi} \vw} \nonumber\\
&\hspace{-2mm} +\frac{\lambda}{2} \norm{\vw}^2 -\frac{1}{p} \sum_{i=1}^{n_t} \ell^\star \left(y_i;u_i \right)+\frac{1}{2} \norm{ \mSigma (\vw - \widehat{\vw}_s) }^2,
\end{align}
where $\vg\in\mathbb{R}^p$ and $\vh\in\mathbb{R}^{n_t}$ are two independent standard Gaussian vectors. The rest of the proof focuses on simplifying the obtained AO formulation and study its asymptotic properties.

\subsubsection{Simplifying the AO Problem of the Target Task}\label{simp_ao_trg}
Here, we focus on simplifying the auxiliary formulation corresponding to the target task. We start our analysis by decomposing the target optimization vector $\vw\in\mathbb{R}^p$ as follows
\begin{align}\label{decom_w_trg}
\vw=({\vxi}_t^\top \vw) {\vxi}_t+\mB_{\vxi_t}^\perp \vr_t.
\end{align}
Here, $\vr_t \in\mathbb{R}^{p-1}$ is a free vector, $\mB_{\vxi_t}^\perp \in\mathbb{R}^{p\times(p-1)}$ is formed by an orthonormal basis orthogonal to the vector $\vxi_t$. Now, define the variable $q_t$ as follows $q_t={\vxi}_t^\top \vw$. Based on the result in Lemma \ref{lem_compact} and the decomposition in \eqref{decom_w_trg}, there exists $C_{q_t}>0$, $C_{r}>0$ and $C_u>0$ such that our auxiliary formulation can be asymptotically expressed in terms of the variables $q_t$ and $\vr_t$ as follows
\begin{align}
&\min_{\substack{(q_t, \vr_t) \in\mathcal{T}_1 }} \max_{\substack{\vu\in\mathcal{C}_t}} \frac{\norm{\vu}}{p} \vg^\top \mB_{\vxi_t}^\perp \vr_t+ \frac{\norm{\vr_t}}{p}  \vh^\top \vu+ \frac{q_t}{p} \vu^\top \vs_{t}+ \frac{\lambda}{2} q_t^2  \nonumber\\
&+\frac{\lambda}{2} \norm{\vr_t}^2 -\frac{1}{p} \sum_{i=1}^{n_t} \ell^\star \left(y_{i};u_i \right)+\frac{1}{2} q_t^2 V_{p,t}- q_t V_{p,ts}\nonumber\\
& + \frac{1}{2} \vr_t^\top ( \mB_{\vxi_t}^\perp )^\top \mLm \mB_{\vxi_t}^\perp \vr_t+ q_t {\vxi}_t^\top \mLm \mB_{\vxi_t}^\perp \vr_t - \vr_t^\top ( \mB_{\vxi_t}^\perp )^\top \mLm \widehat{\vw}_s,\nonumber
\end{align}
Here, we drop terms independent of the optimization variables and the matrix $\mLm\in\mathbb{R}^{p\times p}$ is defined as $\mLm=\mSigma^\top \mSigma$. Additionally, the feasibility set $\mathcal{T}_1$ is defined as follows
\begin{align}\label{T1def}
\mathcal{T}_1=\Big\lbrace (q_t,\vr_t): \abs{q_t}\leq C_{q_t}, \norm{\vr_t}\leq C_r \Big\rbrace.
\end{align}
Here, the sequence of random variables $V_{t,n}$ and $V_{ts,n}$ are defined as follows
\begin{align}
&V_{p,t}={\vxi}_t^\top \mLm {\vxi}_t,~V_{p,ts}={\vxi}_t^\top \mLm \widehat{\vw}_s.
\end{align}
Next, we focus on simplifying the obtained auxiliary formulation. Our strategy is to solve over the direction of the optimization vector $\vr\in\mathbb{R}^{p-1}$. This step requires the interchange of a non--convex minimization and a non--concave maximization. We can easily justify the interchange using the theoretical result in \cite[Lemma A.3]{chris:151}. The main argument is that the strong convexity of the primary formulation in \eqref{po_form_v3} allows us to perform such interchange in the corresponding auxiliary formulation. The optimization problem over the vector $\vr_t$ with fixed norm, i.e. $\norm{\vr_t}=r_t$, can be formulated as follows
\begin{align}\label{form_r}
C_p^\star=&\min_{\vr_t\in\mathbb{R}^{p-1}}~\vb_p^\top \vr_t + \frac{1}{2} \vr_t^\top \mLm^\perp \vr_t,~~\text{s.t.}~~\norm{\vr_t}=r_t,
\end{align}
Here, we ignore constant terms independent of $\vr_t$, the matrix $\mLm^\perp\in\mathbb{R}^{(p-1)\times(p-1)}$ and the vector $\vb_p\in\mathbb{R}^{p-1}$ can be expressed as follows
\begin{align}
\mLm^\perp=( \mB_{\vxi_t}^\perp )^\top \mLm \mB_{\vxi_t}^\perp,~\vb_p&=\frac{\norm{\vu}}{p} (\mB_{\vxi_t}^\perp)^\top \vg + q_t ( \mB_{\vxi_t}^\perp)^\top \mLm {\vxi}_t^\top \nonumber\\
& - ( \mB_{\vxi_t}^\perp )^\top \mLm \widehat{\vw}_s.\nonumber
\end{align}
The optimization problem in \eqref{form_r} is non--convex given the norm equality constraint. It is well--studied in the literature \cite{trust_sp} and is known as the trust region subproblem. Using the same analysis in \cite{dhifallah2020}, the optimal cost value of the optimization problem \eqref{form_r} can be expressed in terms of a one-dimensional optimization problem as follows
\begin{align}
\hspace{-2mm} C_p^\star=\sup_{\sigma>-\mu_p} \left\lbrace -\frac{1}{2} \vb_p^\top [\mLm^\perp+\sigma \mI_{p-1}]^{-1} \vb_p - \frac{\sigma r_t^2}{2} \right\rbrace,
\end{align}
where $\mu_p$ is the minimum eigenvalue of the matrix $\mLm^\perp$, denoted by $\sigma_{\text{min}}(\mLm^\perp)$. This result can be seen by equivalently formulating the non--convex problem in \eqref{form_r} as follows
\begin{align}\nonumber
C_p^\star=&\min_{\vr_t\in\mathbb{R}^{p-1}}\max_{\sigma\in\mathbb{R}}~\vb_p^\top \vr_t + \frac{1}{2} \vr_t^\top \mLm^\perp \vr_t+\frac{\sigma}{2} \left(\norm{\vr_t}^2-r_t^2\right).
\end{align}
Then, show that the optimal $\sigma$ satisfies a constraint that preserves the convexity over $\vw$. This allows us to interchange the maximization and minimization and solve over the vector $\vw$. The above analysis shows that the AO formulation corresponding to our primary problem can be expressed as follows
\begin{align}\label{ao_form_vps}
&\min_{\substack{(q_t, r_t) \in\mathcal{T}_2 }} \max_{\substack{\vu\in\mathcal{C}_t}}\sup_{\sigma>-\mu_p} \frac{r_t}{p} \vh^\top \vu+ \frac{q_t}{p} \vu^\top \vs_{t} + \frac{\lambda}{2} q_t^2 + \frac{\lambda}{2} r_t^2 \nonumber\\
&  -\frac{1}{p} \sum_{i=1}^{n_t} \ell^\star \left(y_{i};u_i \right)+\frac{1}{2} q_t^2 V_{p,t} - q_t V_{p,ts}- \frac{\norm{\vu}^2}{2 p} T_{p,g}(\sigma)\nonumber\\
& -\frac{ \sigma r_t^2}{2}- \frac{1}{2} q_t^2 T_{p,t}(\sigma) -\frac{1}{2} T_{p,s}(\sigma) +  q_t T_{p,ts}(\sigma). 
\end{align}
Here, the set $\mathcal{T}_2$ has the same definition as the set $\mathcal{T}_1$ except that we replace $\norm{\vr_t}$ by $r_t$. Here, the sequence of random functions $T_{p,g}(.)$, $T_{p,t}(.)$, $T_{p,s}(.)$ and $T_{p,ts}(.)$ can be expressed as follows
\begin{align}
\begin{cases}
T_{p,g}(\sigma)= \frac{1}{p} \vg^\top \mB_{\vxi_t}^\perp [\mLm^\perp+\sigma \mI_{p-1}]^{-1} (\mB_{\vxi_t}^\perp)^\top \vg \\
T_{p,t}(\sigma)= {\vxi}_t^\top \mLm \mB_{\vxi_t}^\perp [\mLm^\perp+\sigma \mI_{p-1}]^{-1} (\mB_{\vxi_t}^\perp)^\top \mLm {\vxi}_t \\
T_{p,s}(\sigma)= \widehat{\vw}_s^\top \mLm \mB_{\vxi_t}^\perp [\mLm^\perp+\sigma \mI_{p-1}]^{-1} (\mB_{\vxi_t}^\perp)^\top \mLm \widehat{\vw}_s \\
T_{p,ts}(\sigma)= {\vxi}_t^\top \mLm \mB_{\vxi_t}^\perp [\mLm^\perp+\sigma \mI_{p-1}]^{-1} (\mB_{\vxi_t}^\perp)^\top \mLm \widehat{\vw}_s.
\end{cases}\nonumber
\end{align}
Note that the formulation in \eqref{ao_form_vps} is obtained after dropping terms that converge in probability to zero. This simplification can be justified using a similar analysis as in \cite[Lemma 3]{dhifallah2020}. The main idea is to show that both loss functions converge uniformly to the same limit.

Next, the objective is to simplify the obtained AO formulation over the optimization vector $\vu\in\mathbb{R}^{n_t}$. Based on the property stated in \cite[Lemma 4]{dhifallah2020}, the optimization over the vector $\vu$ can be expressed as follows
\begin{align}
I_{p}^\star&=\max_{\substack{\vu\in\mathcal{C}_t}}r_t \vh^\top \vu+ q_t \vu^\top \vs_{t} -\sum_{i=1}^{n_t} \ell^\star \left(y_{i};u_i \right)- \frac{\norm{\vu}^2}{2} T_{p,g}(\sigma) \nonumber\\
&=\sum_{i=1}^{n_t} \mathcal{M}_{\ell(y_{i},.)}\Big( r_t h_{i}+q_t s_{t,i}; T_{p,g}(\sigma) \Big). \nonumber
\end{align}
This result is valid on events with probability going to one as $p$ goes to $+\infty$. Here, the function $\mathcal{M}_{\ell(y_{i},.)}$ is the Moreau envelope function defined in \eqref{m_env}. The proof of this property is omitted since it follows the same ideas of \cite[Lemma 4]{dhifallah2020}. The main idea is to use Assumption \ref{lossf} to show that the  optimal solution of the unconstrained version of the maximization problem is bounded asymptotically. Then, use the property introduced in  \cite[Example 11.26]{var_prog} to complete the proof. Now, our auxiliary formulation can be asymptotically simplified to a scalar optimization problem as follows
\begin{align}\label{ao_form_v4_tgt}
&\min_{\substack{(q_t, r_t) \in\mathcal{T}_2 }} \sup_{\sigma>-\mu_p}   \frac{\lambda}{2} (q_t^2+r_t^2)  +\frac{1}{2} q_t^2 V_{p,t} -q_t V_{p,ts} -\frac{ \sigma r_t^2}{2}\nonumber\\
&+\frac{1}{p} \sum_{i=1}^{n_t} \mathcal{M}_{\ell(y_{i},.)}\Big( r_t h_{i}+q_ts_{t,i}; T_{p,g}(\sigma) \Big) - \frac{1}{2} q_t^2 T_{p,t}(\sigma)   \nonumber\\
&-\frac{1}{2} T_{p,s}(\sigma)+ q_t T_{p,ts}(\sigma).
\end{align}
Note that the auxiliary formulation in \eqref{ao_form_v4_tgt} has now scalar optimization variables. Then, it remains to study its asymptotic properties. We refer to this problem as the target scalar formulation.

\subsubsection{Asymptotic Analysis of the Target Scalar Formulation} \label{asy_ana_tsf}
In this part, we study the asymptotic properties of the scalar formulation expressed in \eqref{ao_form_v4_tgt}. We start our analysis by studying the asymptotic properties of the sequence of random variables $V_{p,t}$ and $V_{p,ts}$ and the sequence of random functions $T_{p,g}(.)$, $T_{p,t}(.)$, $T_{p,s}(.)$ and $T_{p,ts}(.)$ as given in the following Lemma. 
\begin{lema}[Asymptotic Properties]\label{lemma_cons_srce}
Define $V$ as follows $V=\mathbb{E}_\mu[\mu]$, where the expectation is over the probability distribution $\mathbb{P}_\mu(.)$ defined in Assumption \ref{Sigmaw}. First, the random variable $\mu_p$ converges in probability to $\mu_{\text{min}}$, where $\mu_{\text{min}}$ is defined in Assumption \ref{Sigmaw}. For any fixed $\sigma >0$, the following convergence in probability holds true
\begin{align}
\begin{cases}\nonumber
V_{p,t}\overset{p}{\to} V,~V_{p,ts}\overset{p}{\to}V_{ts}=\rho q_s^\star V\\
T_{p,t}(\sigma-\mu_p)\overset{p}{\to}T_{t}(\sigma-\mu_{\text{min}})\\
T_{p,ts}(\sigma-\mu_p)\overset{p}{\to} T_{ts}(\sigma-\mu_{\text{min}})\\
T_{p,s}(\sigma-\mu_p) \overset{p}{\to} T_s(\sigma-\mu_{\text{min}})\\
T_{p,g}(\sigma-\mu_p) \overset{p}{\to} T_1(\sigma-\mu_{\text{min}}). 
\end{cases}
\end{align}
Here, the deterministic functions $T_t(.)$, $T_{ts}(.)$, $T_s(.)$, $T_1(.)$ and $T_3(.)$ are defined as follows
\begin{align}
\begin{cases}\nonumber
T_{t}(\sigma)=V+\sigma-{1}/T_1(\sigma),~T_{ts}(\sigma)=\rho q_s^\star T_{t}(\sigma)\\
T_s(\sigma)=( (1-\rho^2) (q_s^\star)^2 + (r_s^\star)^2 ) T_3(\sigma)+(\rho q_s^\star)^2 T_{t}(\sigma)\\
T_1(\sigma)={\mathbb{E}_\mu\left[ {1}/{(\mu+\sigma)} \right]},~T_3(\sigma)=\mathbb{E}_\mu\left[ \mu^2/(\mu+\sigma) \right].
\end{cases}
\end{align}
Moreover, the constants $q_s^\star$ and $r_s^\star$ are the optimal solutions of the source asymptotic formulation defined in \eqref{sour_prob_pr}.
\end{lema}
The detailed proof of Lemma \ref{lemma_cons_srce} is provided in Appendix \ref{prf_lemma_cons_srce}. Now that we obtained the asymptotic properties of the sequence of random variables, it remains to study the asymptotic properties of the optimal cost and optimal solution set of the scalar  formulation in \eqref{ao_form_v4_tgt}. To state our first asymptotic result, we define the following deterministic optimization problem
\begin{align}\label{tg_prob_source_pf}
&\min_{\substack{(q_t,r_t)\in\mathcal{T}_2}} \sup_{\sigma>-\mu_{\text{min}}}   \hspace{-1mm}\frac{\lambda}{2} (q_t^2+r_t^2)  +\frac{1}{2} q_t^2 V- q_t V_{ts}-\frac{ \sigma r_t^2}{2} \nonumber\\
&+\alpha_t \mathbb{E}\Big[ \mathcal{M}_{\ell(Y_t,.)}\Big( r_t H_t+q_t S_{t}; T_{g}(\sigma) \Big) \Big]-\frac{1}{2} T_{s}(\sigma) \nonumber\\
&+ q_t T_{ts}(\sigma)- \frac{1}{2} q_t^2 T_{t}(\sigma),
\end{align}
where $H_t$ and $S_t$ are two independent standard Gaussian random variables and $Y_t=\varphi(S_t)$. Here, the function $\mathcal{M}_{\ell(Y_t,.)}$ denotes the Moreau envelope function defined in \eqref{m_env} and the expectation is take over the random variables $H_t$,  $S_t$ and the possibly random function $\varphi(.)$. Now, we are ready to state our asymptotic property of the cost function of \eqref{ao_form_v4_tgt}.
\begin{lema}[Cost Function of the Traget AO Formulation]\label{pconv_tcost}
Define $\mathcal{O}_{p,t}(.)$ as the loss function of the target scalar optimization problem given in \eqref{ao_form_v4_tgt}. Additionally, define $\mathcal{O}_t(.)$ as the cost function of the deterministic formulation in \eqref{tg_prob_source_pf}. Then, the following convergence in probability holds true.
\begin{align}
\mathcal{O}_{p,t}(q_t,r_t,\sigma-\mu_p) \overset{~~p~~}{\to} \mathcal{O}_t(q_t,r_t,\sigma-\mu_{\text{min}}),
\end{align}
for any fixed feasible $q_t$, $r_t$ and $\sigma>0$.
\end{lema}
The proof of the asymptotic property stated in Lemma \ref{pconv_tcost} uses the asymptotic results stated in Lemma \ref{lemma_cons_srce}. Moreover, it uses the weak law of large numbers to show that the empirical mean of the Moreau envelope concentrates around its expected value. Based on Assumption \ref{lossf}, one can see that the following pointwise convergence is valid
\begin{align}
\frac{1}{p} \sum_{i=1}^{n_t} \mathcal{M}_{\ell(y_{i},.)}\big( r_t h_{i}+q_ts_{t,i}; x \big)  \overset{p}{\to} \mathbb{E}\big[\mathcal{M}_{\ell(Y,.)}\big( r_t H+q_t S; x \big) \big].\nonumber
\end{align}
Here $H$ and $S$ are independent standard Gaussian random variables and $Y=\varphi(S)$. The above property is valid for any $x>0$, $r_t\geq 0$ and $q_t$. Based on \cite[Theorem 2.26]{var_prog}, the Moreau envelope function is convex and continuously differentiable with respect $x > 0$. Combining this with \cite[Theorem 7.46]{sto_opt}, the above asymptotic function is continuous in $x>0$. Then, using Lemma \ref{lemma_cons_srce}, the uniform convergence and the continuity property, we conclude that the empirical average of the Moreau envelope converges in probability to the following function
\begin{align}
\mathbb{E}\big[\mathcal{M}_{\ell(Y,.)}\big( r_t H+q_t S; T_g(\sigma-\mu_{\text{min}}) \big) \big],
\end{align}
for any fixed feasible $q_t$, $r_t$ and $\sigma>0$.
This completes the proof of Lemma \ref{pconv_tcost}. The analysis in \cite[Lemma 6]{dhifallah2020} can also be applied here to show that the formulation in \eqref{tg_prob_source_pf} is strictly concave in the maximization variable $\sigma$ for fixed feasible $(q_t,r_t)$. Define the following function
\begin{align}\label{loss_qtrt}
(q_t,r_t) \to \sup_{\sigma>-\mu_{\text{min}}} \mathcal{O}_t(q_t,r_t,\sigma),
\end{align}
where $\mathcal{O}_t(.)$ denotes the cost function of the deterministic formulation in \eqref{tg_prob_source_pf}. The analysis in \cite[Lemma 6]{dhifallah2020} can also be used here to show that the function defined in \eqref{loss_qtrt} is strongly convex in $(q_t,r_t)$ with a strong convexity parameter $\lambda$.

 Now, we use these properties to show that the optimal solution set of the formulation in \eqref{ao_form_v4_tgt} converges in probability to the optimal solution set of the formulation in \eqref{tg_prob_source_pf}.
\begin{lema}[Consistency of the Target AO Formulation]\label{cons_trg}
Define $\mathcal{P}_{p,t}$ and $\mathcal{P}_{t}$ as the optimal set of $(q_t,r_t)$ of the optimization problems formulated in \eqref{ao_form_v4_tgt} and \eqref{tg_prob_source_pf}, respectively. Moreover, define $\mathcal{O}^\star_{p,t}$ and $\mathcal{O}^\star_{t}$ as the optimal cost values of the optimization problems formulated in \eqref{ao_form_v4_tgt} and \eqref{tg_prob_source_pf}, respectively. Then, the following converges in probability holds true
\begin{align}
\mathcal{O}^\star_{p,t} \overset{p}{\to} \mathcal{O}^\star_{t},~\mathbb{D}(\mathcal{P}_{p,t},\mathcal{P}_{t}) \overset{p}{\to} 0,
\end{align}
where $\mathbb{D}(\mathcal{A},\mathcal{B})$ denotes the deviation between the sets $\mathcal{A}$ and $\mathcal{B}$ and is defined as $\mathbb{D}(\mathcal{A},\mathcal{B})=\sup_{\vc_1\in\mathcal{A}}\inf_{\vc_2\in\mathcal{B}} \norm{\vc_1-\vc_2}$.
\end{lema}
The stated result can be proved by first observing that the loss function  $\mathcal{O}_t(.)$ corresponding to the deterministic formulation in \eqref{tg_prob_source_pf} satisfies the following
\begin{align}
\lim_{\sigma \to +\infty} \mathcal{O}_{t}(q_t,r_t,\sigma-\mu_{\text{min}}) = -\infty.
\end{align} 
For any $r_t>0$ and any fixed $q_t$. Combining this with the convergence result in Lemma \ref{pconv_tcost}, \cite[Lemma B.1]{chris:151} and \cite[Lemma B.2]{chris:151}, we obtain the following asymptotic result
\begin{align}
\sup_{\sigma>0} \mathcal{O}_{p,t}(q_t,r_t,\sigma-\mu_p) \overset{~~p~~}{\to} \sup_{\sigma>0} \mathcal{O}_t(q_t,r_t,\sigma-\mu_{\text{min}}).\nonumber
\end{align}
Note that if $r_t=0$, the supremum in the above convergence result occurs at $\sigma\to+\infty$. However, it can be checked that the above convergence result still hold. Based on \cite[Lemma 6]{dhifallah2020}, the cost function of the minimization problem in \eqref{tg_prob_source_pf} is strongly convex in $(q_t,r_t)$. Then, based on \cite[Theorem II.1]{andersen1982} and \cite[Theorem 2.1]{Newey94}, we obtain the convergence result in Lemma \ref{cons_trg}. Now that we obtained the asymptotic problem, it remains to study the asymptotic properties of the training and generalization errors corresponding to the target formulation in \eqref{tform_new}.

\subsubsection{Specialization to the Hard Formulation} 
Before starting the analysis of the generalization error, we specialize our general analysis to the hard transfer formulation. To obtain the asymptotic limit of the hard formulation, we specialize the general results in \eqref{tg_prob_source_pf} to the following probability distribution
\begin{align}\label{hard_prob}
\mathbb{P}_p(\mu)=(1-\delta) d(\mu)+\delta d(\mu-p),
\end{align}
where the function $x \to d(x-a)$ is the dirac delta function defined at $a$. Then, we study the obtained asymptotic results when $p$ goes to $+\infty$. Note that the probability distribution in \eqref{hard_prob} satisfies Assumption \ref{Sigmaw}. Then, the asymptotic limit of the soft formulation corresponding to the probability distribution $\mathbb{P}_\mu(.)$, defined in \eqref{hard_prob}, can be expressed as follows
\begin{align}\label{hard_prob_source_v1}
\min_{\substack{(q_t,r_t)\in\mathcal{T}_2}} &\sup_{\sigma>0} \  \frac{\lambda}{2} (q_t^2+r_t^2) + \frac{T_2(\sigma)}{2} \big( (1-\rho^2) (q_s^\star)^2+(r_s^\star)^2 \big)  \nonumber\\
&\hspace{-5mm}+\alpha_t\mathbb{E}\Big[ \mathcal{M}_{\ell(Y_t,.)}\Big( r_t H_t+q_t S_{t}; T_{1}(\sigma) \Big) \Big]-\frac{\sigma r_t^2}{2} \nonumber\\
&\hspace{-5mm}-\frac{1}{2} \left(q_t-\rho q_s^\star \right)^2 \left(\sigma-1/T_1(\sigma) \right),
\end{align}
where the functions $T_1(.)$ and $T_2(.)$ are defined as follows
\begin{align}
\begin{cases}
T_1(\sigma)=(1-\delta)/\sigma+\delta/(p+\sigma)\\
T_2(\sigma)=\delta p \sigma/(p+\sigma).
\end{cases}
\end{align}
First, one can see that the loss function of \eqref{hard_prob_source_v1}, denoted by $h_p(.)$, converges as follows
\begin{align}
\lim_{p \to +\infty} h_p(q_t,r_t,\sigma)=h(q_t,r_t,\sigma),
\end{align}
for any fixed $q_t$, $r_t$ and $\sigma$ in the feasibility set of the formulation \eqref{hard_prob_source_v1}.
Here, the function $h(.)$ is defined as follows
\begin{align}\label{lossh}
h(q_t,r_t,\sigma)&=\frac{\lambda}{2} (q_t^2+r_t^2) + \frac{\sigma \delta}{2} \left( (1-\rho^2) (q_s^\star)^2+(r_s^\star)^2 \right)  \nonumber\\
&\hspace{-3mm}+\alpha_t\mathbb{E}\Big[ \mathcal{M}_{\ell(Y_t,.)}\Big( r_t H_t+q_t S_{t}; \frac{1-\delta}{\sigma} \Big) \Big]-\frac{\sigma r_t^2}{2} \nonumber\\
&\hspace{-3mm}+\frac{\sigma \delta}{2(1-\delta)} \left(q_t-\rho q_s^\star \right)^2,
\end{align}
for any fixed $q_t$, $r_t$ and $\sigma$ in the feasibility set of the formulation \eqref{hard_prob_source_v1}. 
Based on the analysis in \cite[Lemma 6]{dhifallah2020}, one can see that the formulation in \eqref{hard_prob_source_v1} and the function in \eqref{lossh} are strictly convex in the minimization variables and strictly concave in the maximization variable. Then, based on the analysis in Section \ref{simp_ao_trg} and \cite[Theorem 2.1]{Newey94}, the asymptotic limit of the soft formulation defined in \eqref{hard_prob_source_v1} simplifies to the following formulation
\begin{align}\label{hard_prob_source2}
\min_{\substack{(q_t,r_t)\in\mathcal{T}_2}} \sup_{\sigma>0} \  &\frac{\lambda}{2} (q_t^2+r_t^2) + \frac{\sigma \delta}{2} \left( (1-\rho^2) (q_s^\star)^2+(r_s^\star)^2 \right)  \nonumber\\
&\hspace{-3mm}+\alpha_t\mathbb{E}\Big[ \mathcal{M}_{\ell(Y_t,.)}\Big( r_t H_t+q_t S_{t}; \frac{1-\delta}{\sigma} \Big) \Big]-\frac{\sigma r_t^2}{2} \nonumber\\
&\hspace{-3mm}+\frac{\sigma \delta}{2(1-\delta)} \left(q_t-\rho q_s^\star \right)^2.
\end{align}
This shows that the asymptotic limit of the hard formulation is the deterministic problem \eqref{hard_prob_source2}. 

\subsubsection{Asymptotic Analysis of the Training and Generalization Errors}
First, the generalization error corresponding to the target task is given by
\begin{align}\label{testerr_ana}
{\mathcal{E}}_{\text{test}} &=\frac{1}{4^\upsilon} \mathbb{E}\left[ \left( \varphi(\va_{t,\text{new}}^\top\vxi_t) -\widehat{\varphi}(\widehat{\vw}_t^\top \va_{t,\text{new}}) \right)^2 \right],
\end{align}
where $\va_{t,\text{new}}$ is an unseen target feature vector. Now, consider the following two random variables
\begin{align}
\nu_1=  \va_{t,\text{new}}^\top\vxi_t,~\text{and}~\nu_2=  \widehat{\vw}_t^\top \va_{t,\text{new}}.\nonumber
\end{align}
Given $\widehat{\vw}_t$ and $\vxi_t$, the random variables $\nu_1$ and $\nu_2$ have a bivaraite Gaussian distribution with zero mean vector and covariance matrix given as follows
\begin{align}\label{cov_mtx}
\mC_p=\begin{bmatrix}
\norm{\vxi_t}^2 &  \vxi_t^\top \widehat{\vw}_t \\
  \vxi_t^\top\widehat{\vw}_t &  \norm{\widehat{\vw}_t}^2
\end{bmatrix}.
\end{align}
To precisely analyze the asymptotic behavior of the generalization error, it suffices to analyze the properties of the covariance matrix $\mC_p$. Define the random variables $\widehat{q}_{p,t}^\star$ and $\widehat{r}_{p,t}^\star$ for the target task as follows
\begin{align}\label{opt_po}
&\widehat{q}_{p,t}^\star={\vxi}_t^\top \widehat{\vw}_t,~\text{and}~\widehat{r}_{p,t}^\star=\norm{(\mB_{\vxi_t}^\perp)^\top \widehat{\vw}_t },
\end{align}
where $\mB_{\vxi_t}^\perp$ is defined in Section \ref{simp_ao_trg}. Then, the covariance matrix $\mC_p$ given in \eqref{cov_mtx} can be expressed as follows
\begin{align}
\begin{bmatrix}
1 &  \widehat{q}^{\star}_{p,t} \\
 \widehat{q}^{\star}_{p,t} &  (\widehat{q}^{\star}_{p,t})^2+(\widehat{r}^{\star}_{p,t})^2
\end{bmatrix}.\nonumber
\end{align}
Hence, to study the asymptotic properties of the generalization error, it suffices to study the asymptotic properties of the random quantities $\widehat{q}_{p,t}^\star$ and $\widehat{r}_{p,t}^\star$. 
\begin{lema}[Consistency of the Target Formulation]\label{optm_sl_conv}
The random quantities $\widehat{q}_{p,t}^\star$ and $\widehat{r}_{p,t}^\star$ satisfy the following asymptotic properties
\begin{align}
\widehat{q}^\star_{p,t} \xrightarrow{p} q_t^\star,~\text{and}~\widehat{r}_{p,t}^\star \xrightarrow{p} r_t^\star,\nonumber
\end{align}
where $q_t^\star$ and $r_t^\star$ are the optimal solutions of the deterministic formulation stated in \eqref{tg_prob_source_pf}.
\end{lema}
To prove the above asymptotic result, we define $\widetilde{q}^\star_{p,t}$ and $\widetilde{r}_{p,t}^\star$ as follows
\begin{align}\label{opt_ao}
&\widetilde{q}_{p,t}^\star={\vxi}_t^\top \widetilde{\vw}_t,~\text{and}~\widetilde{r}_{p,t}^\star=\norm{(\mB_{\vxi_t}^\perp)^\top \widetilde{\vw}_t },
\end{align}
where $\widetilde{\vw}_t$ is the optimal solution of the auxiliary formulation in \eqref{ao_form_f}. Given the result in Lemma \ref{cons_trg} and the analysis in Sections \ref{simp_ao_trg} and \ref{asy_ana_tsf}, the convergence result in Lemma \ref{cons_trg} is also satisfied by our auxiliary formulation in \eqref{ao_form_f}, i.e.
\begin{align}
\widetilde{q}^\star_{p,t} \xrightarrow{p} q_t^\star,~\text{and}~\widetilde{r}_{p,t}^\star \xrightarrow{p} r_t^\star.\nonumber
\end{align}
The rest of the proof of the convergence result stated in Lemma \ref{optm_sl_conv} is based on the CGMT framework, i.e. Theorem \ref{mcgmt}. Specifically, it follows after showing that the assumptions in Theorem \ref{mcgmt} are all satisfied. Note that the cost function of the problem \eqref{tg_prob_source_pf} is strongly convex in the minimization variables. Then, based on \cite[Theorem II.1]{andersen1982}, the cost function of the optimization problem in \eqref{ao_form_v4_tgt} converges uniformly to the cost function of \eqref{tg_prob_source_pf}. Combine this with the compactness of the feasibility sets to see that the conditions in Theorem \ref{mcgmt} are all satisfied. Then, the convergence result in Lemma \ref{optm_sl_conv} follows.

Note that the CGMT framework applied to prove Lemma \ref{optm_sl_conv} also shows that the optimal cost value of the soft target formulation in \eqref{tform_new} converges in probability to the optimal cost value of the deterministic formulation given in \eqref{tg_prob_source_pf}. Combining this with the result in Lemma \ref{optm_sl_conv} shows the convergence property of the training error stated in \eqref{trg_conv_soft}. Now, it remains to show the convergence of the generalization error. It suffices to show that the generalization error defined in \eqref{testerr_ana} is continuous in the quantities $\widehat{q}_{p,t}^\star$ and $\widehat{r}_{p,t}^\star$. This follows based on Assumption \ref{fhat} and the continuity under integral sign property \cite{analy}. This shows the convergence result in \eqref{gen_conv_soft} which completes the proof of Theorem \ref{target_soft} and Corollary \ref{corollary_hard}. Note that the above analysis of the soft target formulation in \eqref{tform_new} is valid for any choice of $C_{q_t}$ and $C_{r}$ that satisfy the result in Lemma \ref{lem_compact}. One can ignore these bounds given the convexity properties of the deterministic formulation in \eqref{tg_prob_source_pf}. This leads to the scalar formulations introduced in \eqref{tg_prob_source} and \eqref{hard_prob_source}.

\subsection{Phase Transitions in the Hard formulation}
In this part, we provide a rigorous proof of Proposition \ref{th_phtr} and Proposition \ref{th_phtr_class}. Here, we consider the squared--loss function. In this case, the deterministic source formulation given in \eqref{sc_prob_source} can be simplified as follows
\begin{align}
\label{sc_prob_source_sq}
\min_{ \substack{q_s,r_s\geq 0}}~& \frac{1}{2}\max \Big\lbrace -r_s+\sqrt{\alpha_s}(q_s^2+r_s^2+v_{s}-2q_s c_{s} )^{\frac{1}{2}} ,0 \Big\rbrace^2\nonumber\\
&+\frac{\lambda}{2}(q_s^2+r_s^2).
\end{align}
The constants $v_{s}$ and $c_{s}$ are defined as $v_{s}=\mathbb{E}[Y_s^2]$ and $c_{s}=\mathbb{E}[S_{s} Y_s]$, where $Y_s=\varphi(S_s)$ and $S_s$ is a standard Gaussian random variable. Additionally, the target scalar formulation given in \eqref{tg_prob_source} can be simplified as follows
\begin{align}\label{tg_prob_source_sq}
\min_{\substack{q_t,r_t\geq 0}} \sup_{\sigma>0} \  &\frac{\lambda}{2} (q_t^2+r_t^2) + \frac{\sigma \delta}{2} \left((1-\rho^2) (q_s^\star)^2+(r_s^\star)^2 \right)  \nonumber\\
&\hspace{-3mm}+\frac{\alpha_t \sigma}{2(1-\delta)+2 \sigma} (r_t^2+q_t^2+v_t-2q_t c_t) - \frac{\sigma r_t^2}{2} \nonumber\\
&\hspace{-3mm}+\frac{\sigma \delta}{2(1-\delta)} \left(q_t-\rho q_s^\star \right)^2.
\end{align}
Here, the constants $v_{t}$ and $c_{t}$ are defined as $v_{t}=\mathbb{E}[Y_t^2]$ and $c_{t}=\mathbb{E}[Y_t S_{t}]$, where $Y_t=\varphi(S_t)$ and $S_t$ is a standard Gaussian random variable. Under the conditions stated in Proposition \ref{th_phtr} and Proposition \ref{th_phtr_class}, the source deterministic formulation given in \eqref{sc_prob_source_sq} can be simplified as follows
\begin{align}
\label{sc_prob_source_sq_pf}
\min_{ \substack{q_s,r_s\geq 0}}~&  -r_s+\sqrt{\alpha_s}(q_s^2+r_s^2+v_{s}-2q_s c_{s} )^{\frac{1}{2}}.
\end{align}
Note that one can easily solve over the variables $q_s$ and $r_s$. Specifically, the optimal solutions of \eqref{sc_prob_source_sq_pf} can be expressed as follows
\begin{align}
q_s^\star=c_s,~\text{and}~r_s^\star=\sqrt{v_s-c_s^2}/\sqrt{\alpha_s-1}.
\end{align}
Moreover, the target deterministic formulation given in \eqref{tg_prob_source_sq} can be expressed as follows
\begin{align}\label{hard_prob_source_sq_pf}
\min_{\substack{q_t,r_t\geq 0}} \sup_{\sigma>0} \  &  \frac{\sigma \delta}{2} \beta_2 + \frac{\alpha_t \sigma}{2(1-\delta)+2\sigma} (r_t^2+q_t^2+v_t-2q_t c_t)   \nonumber\\
&-\frac{\sigma r_t^2}{2}+\frac{\sigma \delta}{2(1-\delta)} \left(q_t-\beta_1 \right)^2, 
\end{align}
where $\beta_1$ and $\beta_2$ are given by
\begin{align}
\beta_1=\rho q_s^\star,~\beta_2=\left( (1-\rho^2) (q_s^\star)^2+(r_s^\star)^2 \right).
\end{align}
Before solving the optimization problem in \eqref{hard_prob_source_sq_pf}, we consider the following change of variable
\begin{align}\label{chgvar}
x_t^2=r_t^2-\delta\beta_2-\frac{\delta}{1-\delta}(q_t-\beta_1)^2.
\end{align}
Note that the above change of variable is valid since the formulation in \eqref{hard_prob_source_sq_pf} requires the right hand side of \eqref{chgvar} to be positive. Therefore, the formulation in \eqref{hard_prob_source_sq_pf} can be expressed in terms of $x_t$ instead of $r_t$ as follows
\begin{align}\label{hard_prob_source_sq_pf2}
\min_{\substack{q_t,x_t\geq 0}} \sup_{\sigma>0}  \ &  \frac{\alpha_t \sigma}{2(1-\delta)+2\sigma} \big(x_t^2+\delta\beta_2+\frac{\delta}{1-\delta}(q_t-\beta_1)^2\nonumber\\
&+q_t^2+v_t-2q_t c_t \big)  -\frac{\sigma x_t^2}{2}.
\end{align}
Now, it can be easily checked that the above optimization problem can be solved over the variable $\sigma$ to give the following formulation
\begin{align}
&\min_{\substack{q_t,x_t\geq 0}}   \frac{1}{2} \max\Big\lbrace -x_t \sqrt{1-\delta} + \sqrt{\alpha_t} \big( x_t^2+\delta\beta_2\nonumber\\
&+\frac{\delta}{1-\delta}(q_t-\beta_1)^2+q_t^2+v_t-2q_t c_t \big)^{\frac{1}{2}},0 \Big\rbrace^2.\nonumber
\end{align}
It is now clear that one can solve the problem in \eqref{hard_prob_source_sq_pf2} in closed form. Moreover, it can be easily checked that the optimal solutions of the optimization problem \eqref{hard_prob_source_sq_pf} can be expressed as follows
\begin{align}
\begin{cases}\nonumber
q_t^\star=(1-\delta) c_t + \delta \beta_1\\
(r_t^\star)^2=\frac{1-\delta}{\alpha_t+\delta-1} \left( (\delta-1)c_t^2+\delta \beta_1^2+\delta \beta_2+v_t-2\delta \beta_1c_t \right)\\
~~~~~~~+\delta \beta_2 +\delta(1-\delta)(c_t-\beta_1)^2.
\end{cases}
\end{align}
Then, the asymptotic limit of the generalization error corresponding to the hard formulation can be determined in closed--form. Since the source and target models given in \eqref{smodel} and \eqref{tmodel} use the same data generating function, the constants $v_t$, $c_t$, $v_s$ and $c_s$ are all equal. We express them as $v$ and $c$ in the rest of the proof. 
\subsubsection{Regression Model}
\label{reg_pt_pf}
In this part, we assume that the function $\widehat{\varphi}(.)$ is the identity function. Based on the asymptotic result stated in Corollary \ref{corollary_hard}, the asymptotic limit of the generalization error corresponding to the hard formulation can be expressed as follows
\begin{align}
{\mathcal{E}}_{\text{test}} &=v-2c q_{t}^\star+ (q_t^\star)^2+(r_t^\star)^2.\nonumber
\end{align}
It can be easily checked that the generalization error can be express as follows
\begin{align}
{\mathcal{E}}_{\text{test}}&=\frac{\alpha_t}{\alpha_t+\delta-1} \left( \delta \lbrace (c-\beta_1)^2+\beta_2 \rbrace +(v-c^2)\right).
\end{align}
Note that the the generalization error obtained above depends explicitly on $\delta$. Now, it suffices to study the derivative of ${\mathcal{E}}_{\text{test}}$ to find the properties of the optimal transfer rate $\delta$ that minimizes the generalization error. Note that the derivative can be expressed as follows
\begin{align}\label{ginf}
\hspace{-2mm}{\mathcal{E}}_{\text{test}}^\prime(\delta)=\frac{ (\alpha_t-1)\lbrace (c-\beta_1)^2+\beta_2 \rbrace -(v-c^2) }{(\alpha_t+\delta-1)^2}.
\end{align}
This shows that the derivative of the generalization error has the same sign as the numerator. This means that the optimal transfer rate satisfies the following
\begin{align}\label{ph_tran}
\delta^\star=\begin{cases}
1 & \text{if}~Z_t<0\\
0 & \text{if}~Z_t>0\\
[0~1] & \text{otherwise},
\end{cases}
\end{align}
where $Z_t$ is given by
\begin{align}
Z_t=(\alpha_t-1)\lbrace (c-\beta_1)^2+\beta_2 \rbrace -(v-c^2).
\end{align}
It can be easily shown that the condition in \eqref{ph_tran} can be expressed as the one given in \eqref{asy_delta1}. This completes the proof of Proposition \ref{th_phtr}.
\subsubsection{Classification Model}
\label{class_pt_pf}
In this part, we assume that the function $\widehat{\varphi}(.)$ is the sign function. Based on the asymptotic result stated in Corollary \ref{corollary_hard}, the asymptotic limit of the generalization error corresponding to the hard formulation can be expressed as follows
\begin{align}
{\mathcal{E}}_{\text{test}}&=\frac{1}{\pi} \text{acos}\Big( \frac{q_{t}^\star}{\sqrt{(q_t^\star)^2+(r_t^\star)^2}} \Big).
\end{align}
Given the closed-form expressions of the solutions, the generalization error can be expressed in terms of the transfer rate $\delta$ as follows
\begin{align}
{\mathcal{E}}_{\text{test}}(\delta)&=\frac{1}{\pi} \text{acos}\Big( \frac{(a\delta+c)\sqrt{\delta+\alpha_t-1}}{\sqrt{T_1 \delta^2+T_2 \delta + T_3} } \Big).\nonumber
\end{align}
Here, the constant terms $a$, $T_1$, $T_2$ and $T_3$ are independent of the transfer rate $\delta$ and are given by
\begin{align}
&a=\rho c-c,~T_1=-2 c^2+2c^2 \rho \nonumber\\
&T_2=\alpha_t (v-c^2)/(\alpha_s-1)+4c^2-2c^2 \rho -v\nonumber\\
&T_3=(\alpha_t-2)c^2+v.\nonumber
\end{align}
Given that the $\text{acos}(.)$ function is strictly decreasing, it suffices to find the maximum of the following function
\begin{align}
g(\delta)=\frac{(a\delta+c)\sqrt{\delta+\alpha_t-1}}{\sqrt{T_1 \delta^2+T_2 \delta + T_3} },
\end{align}
to determine the properties of the optimal transfer rate $\delta^\star$ that gives the lowest generalization error. It can be easily checked that the derivative of the function $g(.)$ with respect to $\delta$ can be fully characterized by analyzing the following third degree polynomial 
\begin{align}
h(\delta)=Z_1 \delta^3+Z_2\delta^2+Z_3\delta+Z_4.
\end{align}
Here, $Z_1$, $Z_2$, $Z_3$ and $Z_4$ are independent of $\delta$ and can be expressed as follows
\begin{align}
&Z_1=a T_1,~Z_2=2aT_2-c T_1,\nonumber\\
&Z_3=3aT_3+a(\alpha_t-1)T_2-2c(\alpha_t-1)T_1\nonumber\\
&Z_4=( 2(\alpha_t-1)a+c )T_3-c(\alpha_t-1)T_2.\nonumber
\end{align}
We can see that the function $g(.)$ is increasing at $\delta=0$ when $Z_4> 0$. This means that the function ${\mathcal{E}}_{\text{test}}(.)$ is decreasing at $\delta=0$ when $Z_4> 0$. This means that there exists $\delta_p>0$ such that the hard transfer with $\delta_p$ is better than the standard transfer in this case. It can be easily checked that this is equivalent to the condition provided in Proposition \ref{th_phtr_class}. This completes the proof of the theoretical statement in Proposition \ref{th_phtr_class}.

\section{Conclusion}\label{cond}
In this paper, we presented a precise characterization of the asymptotic properties of two simple transfer learning formulations. Specifically, our results show that the training and generalization errors corresponding to the considered transfer formulations converge to deterministic functions. These functions can be explicitly found by combining the solutions of two deterministic scalar optimization problems. Our simulation results validate our theoretical predictions and reveal the existence of a phase transition phenomenon in the hard transfer formulation. Specifically, it shows that the hard transfer formulation moves from negative transfer to positive transfer when the similarity of the source and target tasks move past a well-defined critical threshold.

\section{Appendix: Proof of Lemma \ref{lemma_cons_srce}}\label{prf_lemma_cons_srce}
To prove the convergence properties stated in Lemma \ref{lemma_cons_srce}, we show first that they are valid for the auxiliary formulation corresponding to the source problem. 
\subsection{Auxiliary Convergence}\label{conv_aux}
Note that the analysis present in Section \ref{tech_deta} is also valid for the source problem. This is because the formulation in \eqref{tform_new} is equivalent to the source problem in \eqref{sform} if $\mSigma$ is the all zero matrix and we use the source training data. Then, we can see that the optimal solution of the auxiliary formulation corresponding to the source problem, denoted by $\widetilde{\vw}_s$, can be expressed as follows
\begin{align}\label{sour_sol}
\widetilde{\vw}_s=q_{p,s}^\star {\vxi}_s- \frac{r_{p,s}^\star}{\norm{\widetilde{\vg}_s}} \mB_{\vxi_s}^\perp \widetilde{\vg}_s,
\end{align}
where $\widetilde{\vg}_s=(\mB_{\vxi_s}^\perp)^\top \vg_s$ and $\vg_s$ has independent standard Gaussian components. Here, $\mB_{\vxi_s}^\perp \in\mathbb{R}^{p\times(p-1)}$ is formed by an orthonormal basis orthogonal to the vector $\vxi_s$. Additionally, our analysis in Section \ref{tech_deta} shows that the following convergence in probability holds
\begin{align}\label{consis_srce}
q_{p,s} \xrightarrow{p} q_s^\star~\text{and}~{r}_{p,s}^\star \xrightarrow{p} r_s^\star.
\end{align}
Here, $q_s^\star$ and $r_s^\star$ are the optimal solutions of asymptotic limit of the source formulation defined in \eqref{sc_prob_source}.

Based on Assumption \ref{Sigmaw}, the random variable $\mu_p$ converges in probability to $\mu_{\text{min}}$, where $\mu_{\text{min}}$ is defined in Assumption \ref{Sigmaw}. Using \cite[Proposition 3]{Dabah}, Assumptions \ref{rad_fv} and \ref{Sigmaw}, the sequence of random variables ${V}_{p,t}$ converges pointwisely in probability to the constant $V=\mathbb{E}_\mu[\mu]$, where the expectation is taken over the probability distribution $\mathbb{P}_\mu(.)$ defined in Assumption \ref{Sigmaw}. Now, we study the properties of the remaining functions using the optimal solution of the auxiliary formulation defined in \eqref{sour_sol}, i.e. $\widetilde{\vw}_s$, instead of $\widehat{\vw}_s$. For instance, we first study the random sequence $\widetilde{V}_{p,ts}={\vxi}_t^\top \mLm \widetilde{\vw}_s$ to infer the asymptotic properties of ${V}_{p,ts}$.

Exploiting the predictions stated in \eqref{sour_sol} and \eqref{consis_srce}, the sequence of random variables $\widetilde{V}_{p,ts}$ converges in probability to the following constant
\begin{align}
&{V}_{ts}=q_s^\star \rho V,
\end{align}
First, fix $\sigma>-\mu_{\text{min}}$. Then, based on the convergence of $\mu_p$ and \cite[Proposition 3]{Dabah}, the sequence of random functions $T_{p,g}(.)$ converges in probability as follows
\begin{align}
T_{p,g}(\sigma) \overset{p}{\to} T_{g}(\sigma)=\mathbb{E}_\mu\left[ {1}/{(\mu+\sigma)} \right].
\end{align}
Now, we express $\sigma$ as $\sigma=\sigma^\prime-x$, where $\sigma^\prime>0$. This means that the following convergence in probability holds true
\begin{align}\label{convTgx}
T_{p,g}(\sigma^\prime-x) \overset{p}{\to} T_{g}(\sigma^\prime-x),
\end{align}
for any $x<\sigma^\prime+\mu_{\text{min}}$. Note that the functions $T_{p,g}(.)$ and $T_g(.)$ are both convex and continuous in the variable $x$ in the set $[0,~\sigma^\prime+\mu_{\text{min}}[$. Then, based on \cite[Theorem II.1]{andersen1982}, the convergence in \eqref{convTgx} is uniform in the variable $x$ in the compact set $[0,~\sigma^\prime/2+\mu_{\text{min}}]$. Now, note that $\mu_p$ converges in probability to $\mu_{\text{min}}$. Therefore, we obtain the following convergence in probability
\begin{align}
T_{p,g}(\sigma^\prime-\mu_p) \overset{p}{\to} T_{g}(\sigma^\prime-\mu_{\text{min}}),
\end{align}
valid for any fixed $\sigma^\prime>0$.
Using the block matrix inversion lemma, the function $T_{p,t}(.)$ can be expressed as follows
\begin{align}
T_{p,t}(\sigma)&= {\vxi}_t^\top \mLm \mB_{\vxi_t}^\perp [ (\mB_{\vxi_t}^\perp)^\top \mLm \mB_{\vxi_t}^\perp +\sigma \mI_{p-1}]^{-1} (\mB_{\vxi_t}^\perp)^\top \mLm {\vxi}_t\nonumber\\
&={\vxi}_t^\top \mLm {\vxi}_t +\sigma - \frac{1}{{\vxi}_t^\top [  \mLm +\sigma \mI_{p}]^{-1} {\vxi}_t}.
\end{align}
Then, using the theoretical results stated in \cite[Proposition 3]{Dabah}, the functions $T_{p,t}(.)$ converges in probability as follows
\begin{align}
T_{p,t}(\sigma) \overset{p}{\to} T_{t}(\sigma)=V+\sigma-\frac{1}{\mathbb{E}_\mu\left[ {1}/{(\mu+\sigma)} \right]}.
\end{align}
Combine this with the above analysis to obtain the following convergence in probability
\begin{align}
T_{p,t}(\sigma^\prime-\mu_p) \overset{p}{\to} T_{t}(\sigma^\prime-\mu_{\text{min}}),
\end{align}
valid for any $\sigma^\prime>0$.
Based on the result in \eqref{sour_sol}, the sequence of random functions $\widetilde{T}_{p,ts}(.)$ converges in probability to the following function
\begin{align}
T_{ts}(\sigma)=q_s^\star \rho T_{t}(\sigma).
\end{align}
Combine this with the above analysis to obtain the following convergence in probability
\begin{align}
\widetilde{T}_{p,ts}(\sigma^\prime-\mu_p) \overset{p}{\to} T_{ts}(\sigma^\prime-\mu_{\text{min}}),
\end{align}
valid for any $\sigma^\prime>0$.
Using the same analysis and based on \eqref{sour_sol} and \eqref{consis_srce}, one can see that the sequence of random functions $\widetilde{T}_{p,s}(.)$ converges in probability to the following function
\begin{align}
\widetilde{T}_{p,s}(\sigma) &\overset{p}{\to} T_s(\sigma)=(\rho q_s^\star)^2 T_{t}(\sigma)\nonumber\\
&+\left( (1-\rho^2) (q_s^\star)^2+(r_s^\star)^2\right) \mathbb{E}_\mu\left[ \mu^2/(\mu+\sigma) \right].
\end{align}
Combine this with the above analysis to obtain the following convergence in probability
\begin{align}
\widetilde{T}_{p,s}(\sigma^\prime-\mu_p) \overset{p}{\to} T_s(\sigma^\prime-\mu_{\text{min}}),
\end{align}
valid for any $\sigma^\prime>0$.
The above analysis shows that the asymptotic properties stated in Lemma \ref{lemma_cons_srce} are valid for the AO formulation corresponding to the source problem. Now, it remains to show that these properties also hold for the primary formulation.
\subsection{Primary Convergence}
Now, we show that the convergence properties proved above are also valid for the primary problem. To this end, we show that all the assumptions in Theorem \ref{mcgmt} are satisfied. We start our proof by defining the following open set
\begin{align}
\mathcal{T}_\epsilon=\lbrace \vw\in\mathbb{R}^p: \abs{{\vxi}_t^\top \mLm {\vw} -V_{ts}} < \epsilon \rbrace.\nonumber
\end{align}
Now, we consider the feasibility set $\mathcal{D}_\epsilon=\mathcal{T}_1/ \mathcal{S}_\epsilon$, where $\mathcal{T}_1$ is defined in \eqref{T1def}. Based on the analysis of the generalized target formulation in Section \ref{simp_ao_trg}, one can see that the AO formulation corresponding to the source formulation with the set $\mathcal{D}_\epsilon$ can be asymptotically expressed as follows
\begin{align}
&\mathfrak{V}_p:\min_{\substack{(q_s, r_s)\in{\mathcal{T}}_2}} \min_{\substack{\vr_s\in\widetilde{\mathcal{D}}_\epsilon}} \max_{\vu\in\mathcal{C}_s}  \frac{\norm{\vu}}{p}  \vg_s^\top \mB_{\vxi_s}^\perp \vr_s+ \frac{q_s}{p} \vu^\top \vs_{s}   \nonumber\\
&+\frac{\lambda}{2} (q_s^2+\norm{\vr_s}^2)+ \frac{1}{p}\norm{\vr_s} \vh_s^\top \vu -\frac{1}{p} \sum_{i=1}^{n_s} \ell^\star \left(y_{s,i};u_i \right).\nonumber
\end{align}
Here, the feasibility set $\mathcal{T}_2$ is defined in Section \ref{simp_ao_trg} and the feasibility set $\widetilde{\mathcal{D}}_\epsilon$ is given by
\begin{align}
&\Big\lbrace \vr_s: \abs{ q_s \rho V_{p,t}+q_s \sqrt{1-\rho^2} V_{p,r}+{\vxi}_t^\top \mLm \mB_{\vxi_s}^\perp \vr_s -V_{ts}} \geq \epsilon\nonumber\\
&~~~~~~~,\norm{\vr_s}=r_s \Big\rbrace.\nonumber
\end{align}
This follows based on the decomposition in \eqref{decom_w_trg} and where $V_{p,t}$  is defined in Section \ref{simp_ao_trg} and $V_{p,r}=\vxi_t^\top \mLm \vxi_r$.
Note that the optimization problem given in $\mathfrak{V}_p$ can be equivalently formulated as follows
\begin{align}
&\mathfrak{V}_p:\min_{\substack{(q_s, r_s)\in \widehat{\mathcal{S}}_\epsilon }} \min_{\substack{\vr_s\in\widetilde{\mathcal{D}}_\epsilon}} \max_{\vu\in\mathcal{C}_s}  \frac{\norm{\vu}}{p}  \vg_s^\top \mB_{\vxi}^\perp \vr_s+ \frac{q_s}{p} \vu^\top \vs_{s}  \nonumber\\
&+\frac{\lambda}{2} (q_s^2+\norm{\vr_s}^2)+ \frac{1}{p}\norm{\vr_s} \vh_s^\top \vu -\frac{1}{p} \sum_{i=1}^{n_s} \ell^\star \left(y_{s,i};u_i \right).\nonumber
\end{align}
Here, we replace the feasibility set $\mathcal{T}_2$ by the feasibility set $\widehat{\mathcal{S}}_\epsilon$ defined as follows
\begin{align}
&\Big\lbrace \abs{ q_s \rho V_{p,t}+q_s \sqrt{1-\rho^2} V_{p,r}-r_s {\vxi}_t^\top \mLm \mB_{\vxi_s}^\perp \frac{\widetilde{\vg}_s}{\norm{\widetilde{\vg}_s}}-V_{ts}}\nonumber\\
& ~~~~~\geq \epsilon \Big\rbrace \cap \mathcal{T}_2,\nonumber
\end{align} 
where $\widetilde{\vg}_s=(\mB_{\vxi_s}^\perp)^\top \vg_s$. This follows since the first set in $\widehat{\mathcal{S}}_\epsilon$ satisfies the condition in the set $\widetilde{\mathcal{D}}_\epsilon$. Now, assume that $\widehat{\phi}_p^\star$ is the optimal cost value of the optimization problem $\mathfrak{V}_p$ and define the function $\widehat{h}_p(.)$ as follows
\begin{align}
&\widehat{h}_p(q_s,r_s)=\min_{\substack{\vr_s\in\widetilde{\mathcal{D}}_\epsilon}} \max_{\vu\in\mathcal{C}_s}  \frac{\norm{\vu}}{p}  \vg_s^\top \mB_{\vxi_s}^\perp \vr_s+ \frac{q_s \vu^\top \vs_{s}}{p}    \nonumber\\
&+\frac{\lambda}{2} (q_s^2+r_s^2)+ \frac{r_s}{p} \vh_s^\top \vu -\frac{1}{p} \sum_{i=1}^{n_s} \ell^\star \left(y_{s,i};u_i \right),\nonumber
\end{align}
in the set $\widehat{\mathcal{S}}_\epsilon$.
Based on the max--min inequality \cite{byd}, the function $\widehat{h}_p(.)$ can be lower bounded by the following function
\begin{align}
&\widetilde{h}_p(q_s,r_s)=\max_{\vu\in\mathcal{C}_s}  \min_{\substack{\vr_s\in\widetilde{\mathcal{D}}_\epsilon}}  \frac{\norm{\vu}}{p}  \vg_s^\top \mB_{\vxi_s}^\perp \vr_s+ \frac{q_s \vu^\top \vs_{s}}{p}   \nonumber\\
&+\frac{\lambda}{2} (q_s^2+r_s^2)+ \frac{r_s}{p} \vh_s^\top \vu -\frac{1}{p} \sum_{i=1}^{n_s} \ell^\star \left(y_{s,i};u_i \right).\nonumber
\end{align}
This is valid for any $(q_s, r_s)\in \widehat{\mathcal{S}}_\epsilon$.
Moreover, note that the following inequality holds true
\begin{align}
\min_{\substack{\vr_s\in\widetilde{\mathcal{D}}_\epsilon}}  \frac{\norm{\vu}}{p}  \vg_s^\top \mB_{\vxi_s}^\perp \vr_s \geq -\frac{\norm{\vu}}{p} \norm{(\mB_{\vxi_s}^\perp)^\top \vg_s} r_s,
\end{align}
for any $(q_s,r_s)\in\widehat{\mathcal{S}}_\epsilon$. Following the generalized analysis in Section \ref{simp_ao_trg}, one can see that the auxiliary problem corresponding to the source formulation can be expressed as follows
\begin{align}\label{ao_form_v2_s}
&\min_{\substack{(q_s,r_s)\in\mathcal{T}_2}} \sup_{\sigma>0} \frac{1}{n_s} \sum_{i=1}^{n_s} \mathcal{M}_{\ell(y_{s,i},.)}\Big( r_s h_{s,i}+q_s s_{s,i}; \frac{r_s \norm{\widetilde{\vg}_s}}{\sqrt{n_s} \sigma}  \Big)  \nonumber\\
&-\frac{r_s \sigma }{2} \frac{\norm{\widetilde{\vg}_s}}{\sqrt{n_s}} + \frac{\lambda}{2} (q_s^2+x_s^2),
\end{align}
This means that the function $\widetilde{h}_p(.)$ can be lower bounded by the cost function of the minimization problem formulated in \eqref{ao_form_v2_s} denoted by $\widehat{g}_{p}(.)$, i.e.
\begin{align}
\widehat{g}_p(q_s,r_s) \leq \widetilde{h}_p(q_s,r_s).
\end{align}
Here, both functions are defined in the feasibility set $\widehat{\mathcal{S}}_\epsilon$. Now, define $\phi_p^\star$ as the optimal cost value of the auxiliary optimization problem corresponding to the source formulation defined in Section \ref{form_ao}. Note that the loss function $\widehat{g}_{p}(.)$ is strongly convex in the variables $(q_s,r_s)$ with strong convexity parameter $\lambda>0$. This means that for any $\beta\in[0,1]$, $(q_{s,1},r_{s,1})\in{\mathcal{T}_2}$ and $(q_{s,2},r_{s,2})\in{\mathcal{T}_2}$, we have the following inequality
\begin{align}\label{ineq1}
&\widehat{g}_{p}(\beta\vv_1+(1-\beta)\vv_2)\leq \beta \widehat{g}_{p}(\vv_1)\nonumber\\
&+(1-\beta)\widehat{g}_{p}(\vv_2)-\frac{\lambda}{2} \beta (1-\beta)\norm{\vv_1-\vv_2}^2,
\end{align}
where $\vv_1=[q_{s,1},r_{s,1}]$ and $\vv_2=[q_{s,2},r_{s,2}]$.
Take $\vv_1$ as $\vv_{p}^\star$ which represents the optimal solution of the optimization problem \eqref{ao_form_v2_s}. Then, the inequality in \eqref{ineq1} implies the following inequality
\begin{align}\label{ineq2}
&\phi_p^\star \leq \widehat{g}_{p}(\vv_2)-\frac{\lambda}{2} \beta \norm{\vv_p^\star-\vv_2}^2.
\end{align}
This is valid for any $\vv_2$ in the set $\mathcal{T}_2$.
Now, taking $\beta=1/2$ and the minimum over $\vv_2$ in the set $\widehat{\mathcal{S}}_\epsilon$ in both sides, we obtain the following inequality
\begin{align}\nonumber
\phi_p^\star+\frac{\lambda}{4} \min_{\vv \in \widehat{\mathcal{S}}_\epsilon}\norm{\vv_{p}^\star-\vv}^2 \leq \min_{\vv\in\widehat{\mathcal{S}}_\epsilon}\widehat{g}_{p}(\vv).
\end{align}
Based on the above analysis, note that the following inequality also holds true
\begin{align}
\min_{\vv\in\widehat{\mathcal{S}}_\epsilon}\widehat{g}_{p}(\vv) \leq \widehat{\phi}_p^\star.
\end{align}
Then, to verify the assumption of \cite[Theorem 6.1]{chris:151}, it remains to show that there exists $\epsilon^{\prime}>0$ such that, the following inequality holds
\begin{align}\label{wtprove}
\frac{\lambda}{4} \min_{\vv\in\widehat{\mathcal{S}}_\epsilon}\norm{\vv_{p}^\star-\vv}^2 \geq \epsilon^{\prime},
\end{align}
with probability going to $1$ as $p\to\infty$. Note that any element in the set $\widehat{\mathcal{S}}_\epsilon$ satisfies the following inequality
\begin{align}
&\epsilon \leq \abs{ q_s \rho V_{p,t}+q_s \sqrt{1-\rho^2} V_{p,r}-r_s {\vxi}_t^\top \mLm \mB_{\vxi}^\perp \frac{\widetilde{\vg}_s}{\norm{\widetilde{\vg}_s}} - V_{ts}} \leq \nonumber\\
&\abs{q_s \rho V_{p,t}-V_{ts}}+\abs{q_s \sqrt{1-\rho^2}} \abs{V_{p,r}}+ \abs{r_s} \abs{ {\vxi}_t^\top \mLm \mB_{\vxi}^\perp \frac{\widetilde{\vg}_s}{\norm{\widetilde{\vg}_s}} }\nonumber.
\end{align}
Based on the analysis in Section \ref{conv_aux}, we have the following convergence in probability 
\begin{align}
\abs{q_s \rho V_{p,t}-V_{ts}} \overset{p}{\to} \abs{q_s-q_s^\star} \rho V\nonumber\\
\abs{q_s} \sqrt{1-\rho^2} \abs{V_{p,r}} \overset{p}{\to} 0,~\abs{r_s} \abs{ {\vxi}_t^\top \mLm \mB_{\vxi}^\perp \frac{\widetilde{\vg}_s}{\norm{\widetilde{\vg}_s}} } \overset{p}{\to} 0.
\end{align}
This means that there exists $\epsilon^{\prime \prime}>0$ such that any elements in the set $\widehat{\mathcal{S}}_\epsilon$ satisfies the following inequality
\begin{align}
\abs{q_s-q_s^\star} \rho V \geq \epsilon^{\prime \prime},
\end{align}
with probability going to $1$ as $p\to\infty$. Combining this with Assumption \ref{Sigmaw} and the consistency result stated in \eqref{consis_srce} shows that there exists $\epsilon^{\prime}>0$ such that the following inequality holds
\begin{align}
\frac{\lambda}{4} \min_{\vv\in\widehat{\mathcal{D}}_\epsilon}\norm{\vv_{p}^\star-\vv}^2 \geq \epsilon^{\prime},
\end{align}
with probability going to $1$ as $p\to\infty$. This also proves that there exists $\epsilon^{\prime}>0$ such that the following inequality holds
\begin{align}
\widehat{\phi}_p^\star \geq  \phi_p^\star+\epsilon^{\prime},
\end{align}
with probability going to $1$ as $p\to\infty$. This completes the verification of the assumptions in Theorem \ref{mcgmt}. This means that the optimal solution of the primary problem belongs to the set $\mathcal{S}_\epsilon$ on events with probability going to $1$ as $p\to\infty$. Since the choice of $\epsilon$ is arbitrary, we obtain the following asymptotic result
\begin{align}
V_{p,ts}={\vxi}_t^\top \mLm \widehat{\vw}_s \overset{p}{\to} q_s^\star \rho V,
\end{align}
where $\widehat{\vw}_s$ is the optimal solution of the source problem \eqref{sform}. Following the same analysis, one can also show the remaining convergence properties stated in Lemma \ref{lemma_cons_srce}.

\balance
\bibliographystyle{IEEEtran}
\bibliography{refs}

\end{document}